\documentclass[runningheads]{llncs}

\usepackage{eccv}

\usepackage{eccvabbrv}

\usepackage{graphicx}
\usepackage{booktabs}
\usepackage{multirow}
\usepackage{array}

\usepackage[accsupp]{axessibility}  

\usepackage{graphicx}
\usepackage{booktabs}
\usepackage[table]{xcolor}
\definecolor{eccvblue}{rgb}{0.12,0.49,0.85}
\definecolor{contamred}{rgb}{0.78, 0.16, 0.16}
\usepackage{hyperref}

\usepackage{orcidlink}

\begin{document}

\title{SAM+D: Parameter-Efficient Dimensional Lifting of SAM-Family Models via Depth-Routed LoRA and Depth Shifting} 

\titlerunning{SAM+D}

\author{Yu Song\inst{1}$^{,\star}$\orcidlink{0000-0002-4660-5430} \and
Hao Sun\inst{1}$^{,\star}$\orcidlink{0000-0001-8094-1991} \and
Shiyu Teng\inst{1}\orcidlink{0000-0001-7233-5027} \and
Ikuko Nishikawa\inst{1}\orcidlink{0000-0003-4780-0155} \and
Yen-wei Chen\inst{1}$^{,\dagger}$\orcidlink{0000-0002-5952-0188}}

\authorrunning{Y.~Song et al.}

\institute{
College of Information Science and Engineering, Ritsumeikan University\\
2-150 Iwakura-cho, Ibaraki, Osaka 567-8570, Japan\\
\email{yusong@fc.ritsumei.ac.jp, chen@is.ritsumei.ac.jp}\\
$^\star$~Equal contribution.\quad$^\dagger$~Corresponding author.
}

\maketitle

\begin{abstract}
Existing methods for adapting 2D foundation models such as SAM to 3D volumes either process slices independently---ignoring inter-slice context---or require substantial architectural changes and retraining. In this paper, we present \textbf{SAM+D}, a parameter-efficient framework that lifts SAM-family models by one spatial dimension---enabling 3D volumetric segmentation from 2D SAM and, for the first time via parameter-efficient fine-tuning, end-to-end 4D (3D+T) spatiotemporal segmentation from video-based SAM2---while keeping the vast majority of pre-trained parameters frozen. SAM+D introduces two lightweight, model-agnostic modules into frozen transformer blocks: (1)~\textbf{Depth-Routed LoRA (DRLoRA)} experts with learned routing for spatially adaptive low-rank updates, and (2)~\textbf{Depth Shift Modules (DSM)} for cross-slice feature exchange at zero additional parameter cost. Together, they provide volume-level context while tuning only
${\sim}$2.8\% of parameters for SAM and ${\sim}$3.7\% for SAM2. We evaluate SAM+D in two distinct settings, each lifting the base model by one spatial dimension: 3D segmentation, where SAM
(2D$\,\to\,$3D) is evaluated on four CT benchmarks (KiTS, Pancreas, LiTS, Colon), and 4D segmentation, where SAM2 (2D+T$\,\to\,$3D+T) is evaluated on a cell tracking challenge (CTC) dataset (Fluo-N3DH-SIM+). In both settings SAM+D achieves competitive or superior results under the single-point prompt setting while using fewer trainable parameters than existing methods, demonstrating that SAM+D generalizes across SAM-family architectures, target dimensionalities (3D, 4D), and domains spanning medical imaging and bio-scene understanding. Code is publicly available at \url{https://github.com/JerrySongCST/SAM-Plus-D}.
\end{abstract}

\section{Introduction}
\label{sec:intro}

In recent years, a large number of vision foundation models have focused on 2D images, from self-supervised feature extractors such as VAEs~\cite{kingma2013auto} and DINO~\cite{oquab2023dinov2}, to diffusion-based image generators like stable diffusion~\cite{rombach2022high}, to segmentation foundation models such as the Segment Anything Model (SAM) family~\cite{kirillov2023segment,ravi2024sam, carion2025sam}. However, in a wide range of real-world applications—from CT tumor segmentation for medical diagnosis to cell tracking on time-lapse microscopic images—current 2D models substantially lack the depth context that distinguishes overlapping structures. Take SAM~\cite{kirillov2023segment} as an example: it is trained on over one billion masks on 2D images, while SAM2~\cite{ravi2024sam} extends it to 2D video via tracklet-based memory propagation, but neither provides volumetric understanding. This gap has fuelled urgent interest in spatial intelligence—equipping foundation models with genuine volumetric understanding in 3D and 4D (3D+T).

Volumetric segmentation has traditionally relied on fully 3D architectures trained from scratch~\cite{cciccek20163d,isensee2021nnu}, which model inter-slice context natively but require large annotated
3D datasets that are hard to obtain in many domains, especially in medical and biomedical domains, where dataset sizes are relatively small. Recent 3D segmentation foundation models~\cite{wang2025sam,du2024segvol,ma2025medsam2} scale this paradigm by rebuilding or fully fine-tuning SAM/SAM2 with 3D
components on massive datasets, where 2D priors are discarded in the process. This motivates a second line of work that directly adapts pre-trained 2D model weights for volumetric input, ranging from
injecting 3D adapters or attention layers into the frozen backbone~\cite{chen2024ma,wu2025medical,zhuang2025bio2vol}, to reinterpreting volumes as spatial videos for SAM2~\cite{yang2025sam2,zhu2024medical}, to inserting minimal spatial adapters with very few trainable parameters~\cite{gong20233dsam}.

Despite the recent progress, existing methods share several limitations. First, existing adapters like Low-Rank Adaptation (LoRA)~\cite{hu2022lora} apply the same transformation everywhere in the volume, ignoring that features such as sharp boundaries and homogeneous interiors need different
adaptations---position-aware routing is missing. Second, modelling cross-slice dependencies still requires 3D convolutions or attention that add extra parameters and memory; no method achieves inter-slice communication at zero cost. Third, each method targets a single base model---either SAM or SAM2---so every new foundation model needs a new adapter design; no unified architecture lifts both SAM to 3D and SAM2 to 4D with the same design.

To address the limitations mentioned above, we introduce "SAM+D" in this paper, which introduces two low-cost, parameter-efficient modules into each frozen pre-trained transformer encoder block to lift SAM-family models by one spatial dimension. The first, Depth-Routed LoRA (DRLoRA), maintains multiple LoRA experts per block and routes among them based on the slice's depth position, so that boundary regions and organ interiors can receive distinct low-rank updates. The second, Depth Shift Module (DSM) repurposes the Temporal Shift Module (TSM)~\cite{lin2019tsm}, originally designed for video CNNs, to shift a fraction of channels between neighbouring slices before self-attention---giving each token access to adjacent-slice features at zero extra parameters or FLOPs. By resampling the entire volume to a fixed depth and processing all slices jointly in a single forward pass, SAM+D operates as a patch-free method~\cite{wang2021patch, jeon2025no}, preserving global spatial context without sliding window inference. Neither module assumes anything about the base architecture beyond standard transformer blocks; the same pair therefore lifts SAM (2D$\to$3D) and SAM2 (2D+T$\to$3D+T) without redesign.

Our contributions can be summarized as follows:
\begin{itemize}
\item We propose SAM+D, a unified parameter-efficient framework that lifts SAM-family models by one spatial dimension---enabling 3D volumetric segmentation from SAM and 3D+T spatiotemporal segmentation
from SAM2---using the same lightweight components while keeping the vast majority of pre-trained parameters frozen.
\item We introduce Depth-Routed LoRA (DRLoRA), which learns to blend multiple low-rank experts based on spatial context, and Depth Shift Modules (DSM) that shift a fraction of feature channels between adjacent slices before self-attention, accumulating cross-slice interaction across layers at zero cost (zero added parameters, zero multiply--accumulate (MAC) operations).
\item We evaluate SAM+D on both SAM and SAM2, covering 3D and 3D+T settings across medical imaging and cell tracking tasks, achieving competitive or superior results on five benchmarks---four CT organ segmentation tasks (KiTS~\cite{heller2021state}, Pancreas~\cite{antonelli2022medical}, LiTS~\cite{bilic2023liver}, Colon~\cite{antonelli2022medical}) and one Cell Tracking Challenge (CTC)
data (Fluo-N3DH-SIM+)~\cite{ulman2017objective}---while training only ${\sim}$2.8\% and ${\sim}$3.7\% of total parameters.
\end{itemize}

\section{Related Work}
\subsection{SAM-Family Models}
\label{sec:rw_sam}

The Segment Anything Model (SAM)~\cite{kirillov2023segment} introduced a promptable segmentation paradigm trained on over 11~million images and one billion masks. Its ViT-based image encoder \cite{dosovitskiy2020image}, combined with a lightweight prompt encoder and mask decoder, delivers strong zero-shot segmentation across diverse 2D domains. SAM2~\cite{ravi2024sam} extends this paradigm to video by replacing the ViT encoder with a hierarchical Hiera backbone~\cite{ryali2023hiera} and introducing a streaming memory architecture---comprising a memory encoder, memory bank, and memory attention module---that propagates object identity across frames via tracklet-based matching.  More recently, SAM3~\cite{carion2025sam} further extends the SAM family with concept-based segmentation, introducing a text encoder and visual detector that enable segmentation of objects specified by text queries or visual exemplars within video tracklets.

Despite their architectural differences, all SAM-family models share a fundamental limitation: they operate natively in two spatial dimensions (2D for SAM, 2D+T for SAM2). Volumetric data---3D medical scans, 3D+T time-lapse microscopy---cannot be processed without either discarding inter-slice context (slice-by-slice application) or modifying the architecture. SAM+D addresses this limitation by lifting any SAM-family models by one spatial dimension through lightweight, model-agnostic insertions.

\subsection{Parameter-Efficient Fine-Tuning}
\label{sec:rw_peft}

Parameter-efficient fine-tuning (PEFT) adapts large pretrained models by updating only a small fraction of parameters.  Early strategies include prompt tuning~\cite{lester2021prompt} and prefix tuning~\cite{li2021prefix}, which prepend learnable tokens to the input sequence, and adapter modules~\cite{houlsby2019parameter,chen2022adaptformer}, which insert lightweight bottleneck layers into transformer blocks.  Visual Prompt Tuning (VPT)~\cite{jia2022vpt} extends prompt tuning to vision transformers by injecting learnable tokens into each layer's input space.  Low-Rank Adaptation (LoRA)~\cite{hu2022lora} has since become the dominant paradigm, injecting trainable low-rank matrices into frozen linear projections with minimal inference overhead, and has been widely adopted for both language and vision models~\cite{zhang2023customized,hu2024lga}.  Subsequent works refine LoRA along several axes: QLoRA~\cite{dettmers2023qlora} combines 4-bit quantisation with LoRA for memory-efficient fine-tuning, DoRA~\cite{liu2024dora} decomposes weight updates into magnitude and direction components, and AdaLoRA~\cite{zhang2023adalora} adaptively allocates rank budgets across layers based on importance scores.  Several works further extend LoRA with mixture-of-experts (MoE) routing---MoLoRA~\cite{zadouri2023pushing}, MoLE~\cite{wu2024mixture}, and MixLoRA~\cite{li2024mixlora}---training multiple LoRA experts with learned gating to improve multi-task generalisation. 

However, all existing MoE-LoRA methods route on task identity or token content and require auxiliary load-balancing or importance losses to avoid expert collapse. DRLoRA is categorically different: it routes on the normalised depth position $z$, which makes balancing losses unnecessary and shrinks the router to ${\sim}0.14$K parameters (vs.\ ${\sim}3$--$6$K for content routers).

\subsection{Adapting Foundation Models for Volumetric Segmentation}
\label{sec:rw_volumetric}

Existing approaches to volumetric segmentation span a wide spectrum. At one end, fully 3D architectures such as nnU-Net~\cite{isensee2021nnu} and dedicated 3D foundation models (SAM-Med3D~\cite{wang2025sam}, SegVol~\cite{du2024segvol}, MedSAM2~\cite{ma2025medsam2}) model inter-slice context natively but forfeit 2D pretrained priors or require massive volumetric corpora. At the other end, adapter-based methods retain a frozen 2D backbone: 3DSAM-adapter~\cite{gong20233dsam} inserts depthwise 3D convolutions, MA-SAM~\cite{chen2024ma} uses factorised 3D adapters, and Med-SA~\cite{wu2025medical} bifurcates attention into spatial and depth branches (SD-Trans)---the closest prior work to our proposed DSM, though SD-Trans doubles the attention computation whereas DSM incurs zero additional cost. More recently, SAM2-based methods~\cite{yang2025sam2,zhu2024medical} treat 3D volumes as spatial videos, but remain SAM2-specific and address 3D segmentation only.

All existing approaches target a single base model and a single dimensionality; none provides a unified mechanism that lifts both SAM (2D$\,\to\,$3D) and SAM2 (2D+T$\,\to\,$3D+T) with same design.

\section{Proposed Method}
\label{sec:method}

\subsection{Overview}

\begin{figure}[tb]
  \centering
  \includegraphics[width=\linewidth]{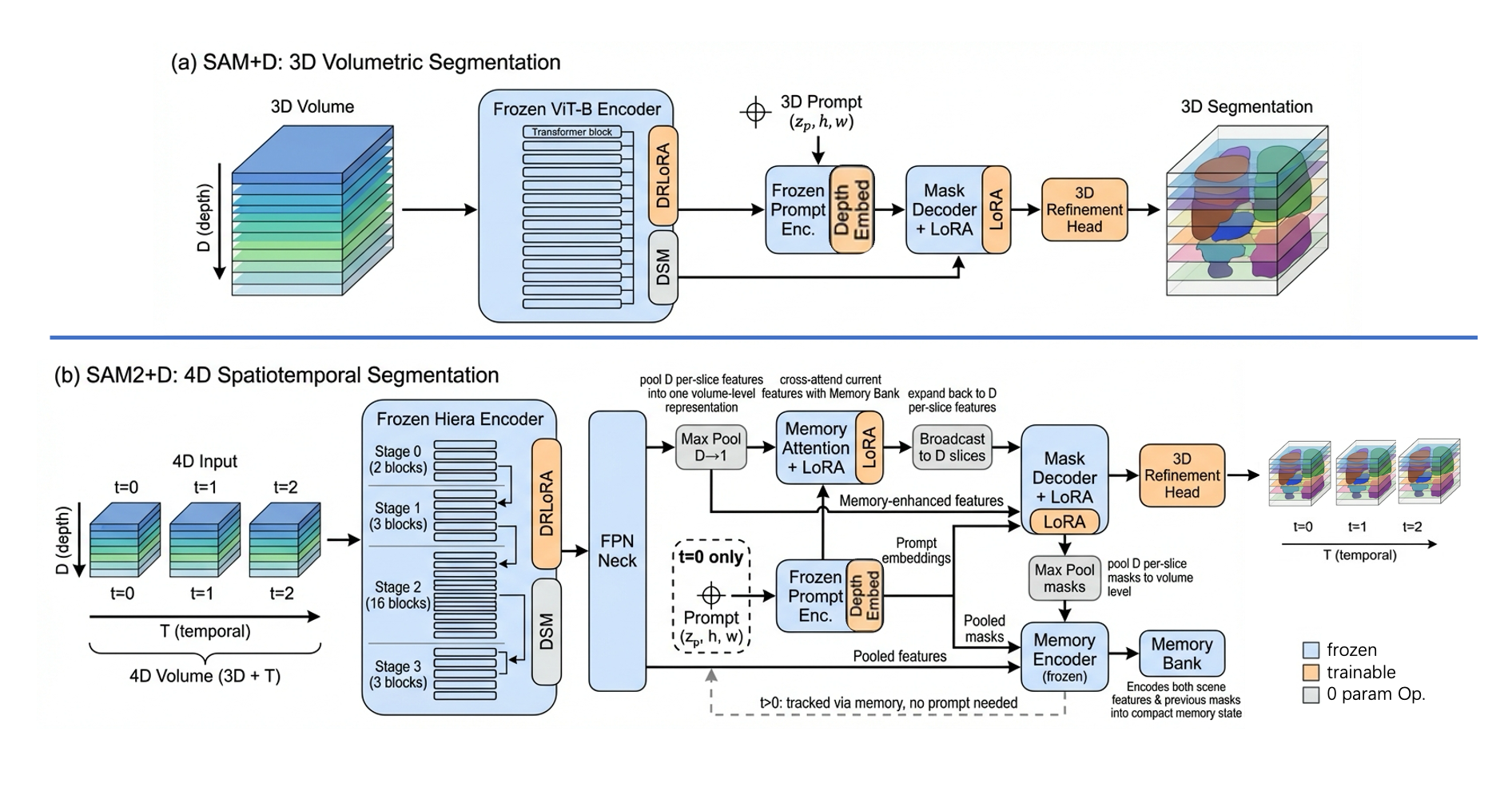}
 \caption{%
  \textbf{Overview of SAM+D.}
  (a)~3D setting: DRLoRA and DSM are inserted into SAM's frozen ViT-B encoder; a LoRA-adapted decoder produces masks refined by a 3D head.
  (b)~4D setting: the same modules are inserted into SAM2's frozen Hiera encoder; depth-pooled features interface with LoRA-adapted memory attention for prompt-free tracking at $t>0$.
  Blue: frozen; orange: trainable; gray: zero-parameter operation.
}
  \label{fig:overview}
\end{figure}

SAM+D lifts frozen SAM-family models by one spatial dimension through two lightweight modules inserted into every transformer block, as shown in Fig.~\ref{fig:overview}. Depth-Routed LoRA Experts (DRLoRA) replaces the standard single-LoRA adaptation with multiple low-rank experts whose contributions are blended by a learned router conditioned on depth position, capturing the structural variation that arises along the depth axis of volumetric data (\S\ref{sec:drlora}). Depth Shift Modules (DSM) exchange a fraction of feature channels between adjacent slices before each attention computation, providing cross-slice context at zero additional parameters or FLOPs (\S\ref{sec:dsm}).

We instantiate SAM+D in two settings. In the 3D setting (\S\ref{sec:3dsam}), DRLoRA and DSM are inserted into SAM's frozen ViT-B encoder, lifting 2D image segmentation to 3D volumetric segmentation of CT scans. In the 4D setting (\S\ref{sec:3dsam2}), the same modules are inserted into SAM2's hierarchical Hiera encoder, lifting 2D+T video segmentation to 3D+T spatiotemporal segmentation, with SAM2's native memory mechanism handling temporal propagation. In both cases, only around ${\sim}$2.8\% and ${\sim}$3.7\% of total parameters are trainable; all pretrained encoder weights remain frozen.

\subsection{Depth-Routed LoRA Experts (DRLoRA)}
\label{sec:drlora}
\begin{figure}[tb]
  \centering
  \includegraphics[width=\linewidth]{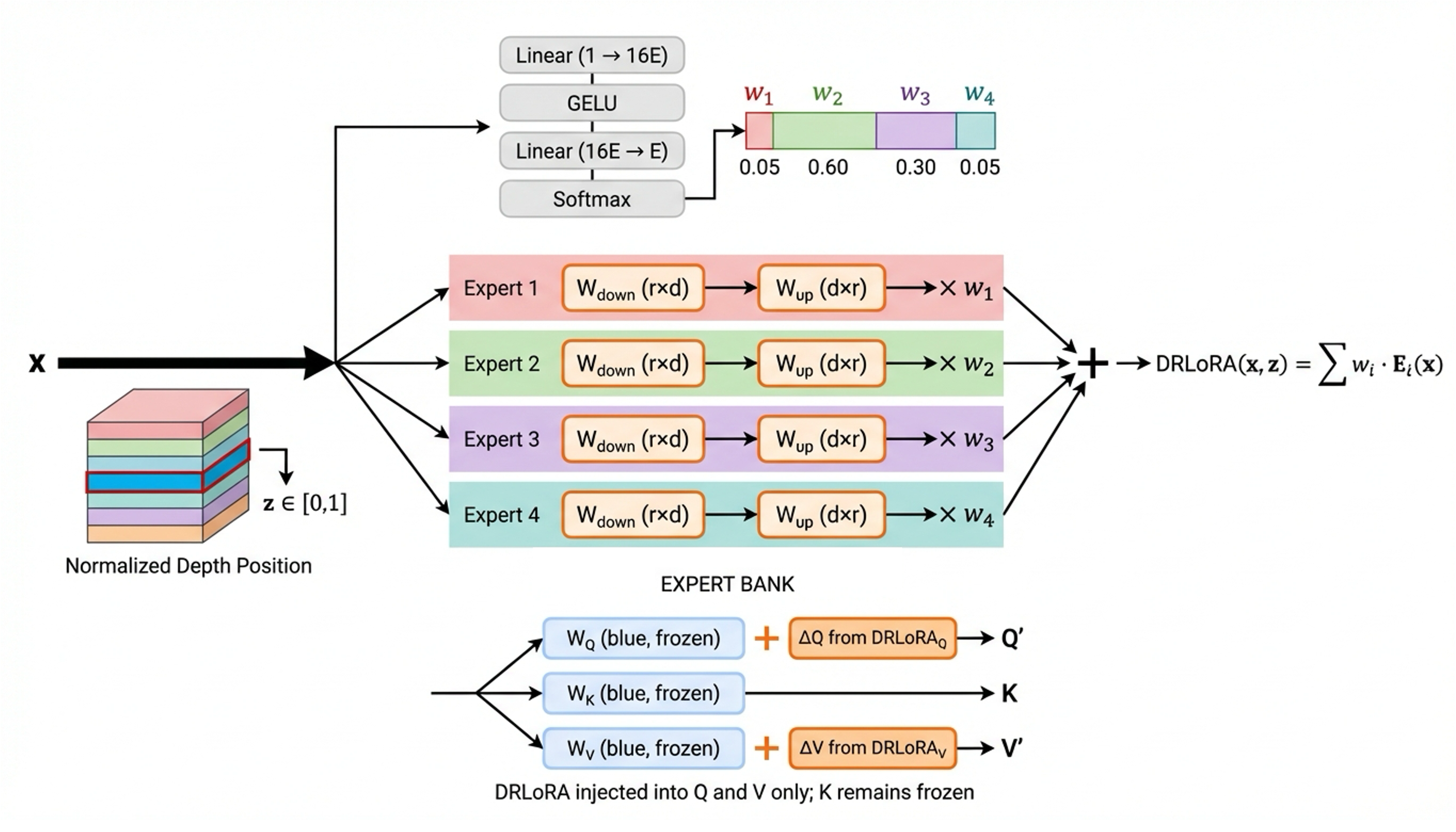}
  \caption{Depth-Routed LoRA (DRLoRA): a lightweight router maps the
  normalised depth position $z$ to a softmax distribution over $E$
  LoRA experts. The weighted sum of expert outputs is added to the
  frozen Q and V projections, producing depth-dependent adaptations
  while leaving K unchanged.}
  \label{fig:drlora}
\end{figure}

LoRA~\cite{hu2022lora} injects trainable low-rank matrices into frozen layers, providing an efficient
alternative to full fine-tuning.  However, a single LoRA applies the same update to every slice, ignoring the fact that adaptation needs vary systematically along the depth axis---e.g., anatomy near the liver dome differs markedly from pelvic slices. DRLoRA addresses this with a mixture-of-experts formulation that conditions the adaptation on depth position, as shown in Fig.~\ref{fig:drlora}.

Concretely, DRLoRA maintains $E$ parallel LoRA experts
$\{\mathcal{E}_1, \ldots, \mathcal{E}_E\}$, each a standard low-rank
factorisation:
\begin{equation}
  \mathcal{E}_i(\mathbf{x})
  = \mathbf{W}_{\mathrm{up}}^{(i)}\,
    \mathbf{W}_{\mathrm{down}}^{(i)}\,\mathbf{x},
  \quad
  \mathbf{W}_{\mathrm{down}}^{(i)} \!\in\! \mathbb{R}^{r \times d},\;
  \mathbf{W}_{\mathrm{up}}^{(i)} \!\in\! \mathbb{R}^{d \times r},
\end{equation}
where $d$ is the embedding dimension and $r \!\ll\! d$ is the rank.
$\mathbf{W}_{\mathrm{up}}^{(i)}$ is initialised to zero so that
training begins from the identity.  To select among experts, a
lightweight MLP routes based on the normalised depth position
$z = d/D_s \in [0,1]$:
\begin{equation}
  \mathbf{w}
  = \mathrm{softmax}\!\bigl(\mathcal{R}(z)\bigr),
  \quad
  \mathcal{R}\!: \mathbb{R}^{1}
    \!\xrightarrow{\text{Lin}}\! \mathbb{R}^{h}
    \!\xrightarrow{\text{GELU}}\! \mathbb{R}^{h}
    \!\xrightarrow{\text{Lin}}\! \mathbb{R}^{E},
\end{equation}
where $h$ is the hidden dimension.  The final output is the
soft-weighted combination of all experts:
\begin{equation}
  \mathrm{DRLoRA}(\mathbf{x}, z)
  = \sum_{i=1}^{E} w_i \cdot \mathcal{E}_i(\mathbf{x}).
  \label{eq:drlora}
\end{equation}
We inject DRLoRA into the \emph{query} and \emph{value} projections
of each self-attention layer, leaving keys frozen:
\begin{equation}
  \mathbf{Q}' = \mathbf{W}_Q\mathbf{x}
    + \mathrm{DRLoRA}_Q(\mathbf{x}, z),
  \quad
  \mathbf{V}' = \mathbf{W}_V\mathbf{x}
    + \mathrm{DRLoRA}_V(\mathbf{x}, z),
  \quad
  \mathbf{K}  = \mathbf{W}_K\mathbf{x}.
\end{equation}
Following the finding of Hu et al.~\cite{hu2022lora} that adapting
both $W_Q$ and $W_V$ outperforms other combinations---including
higher-rank adaptation of a single matrix---under a fixed parameter
budget, we apply DRLoRA to these two projections.  Intuitively,
adapting queries modifies the attention pattern---which tokens
each position attends to---while adapting values modifies the
information content that is propagated through attention.
Together they give DRLoRA control over both where and
what the model attends to in a depth-dependent manner,
while the frozen keys preserve the pre-trained feature space as a
stable matching reference.

\subsection{Depth Shift Module (DSM)}
\label{sec:dsm}

\begin{figure}[tb]
  \centering
  \includegraphics[width=0.7\linewidth]{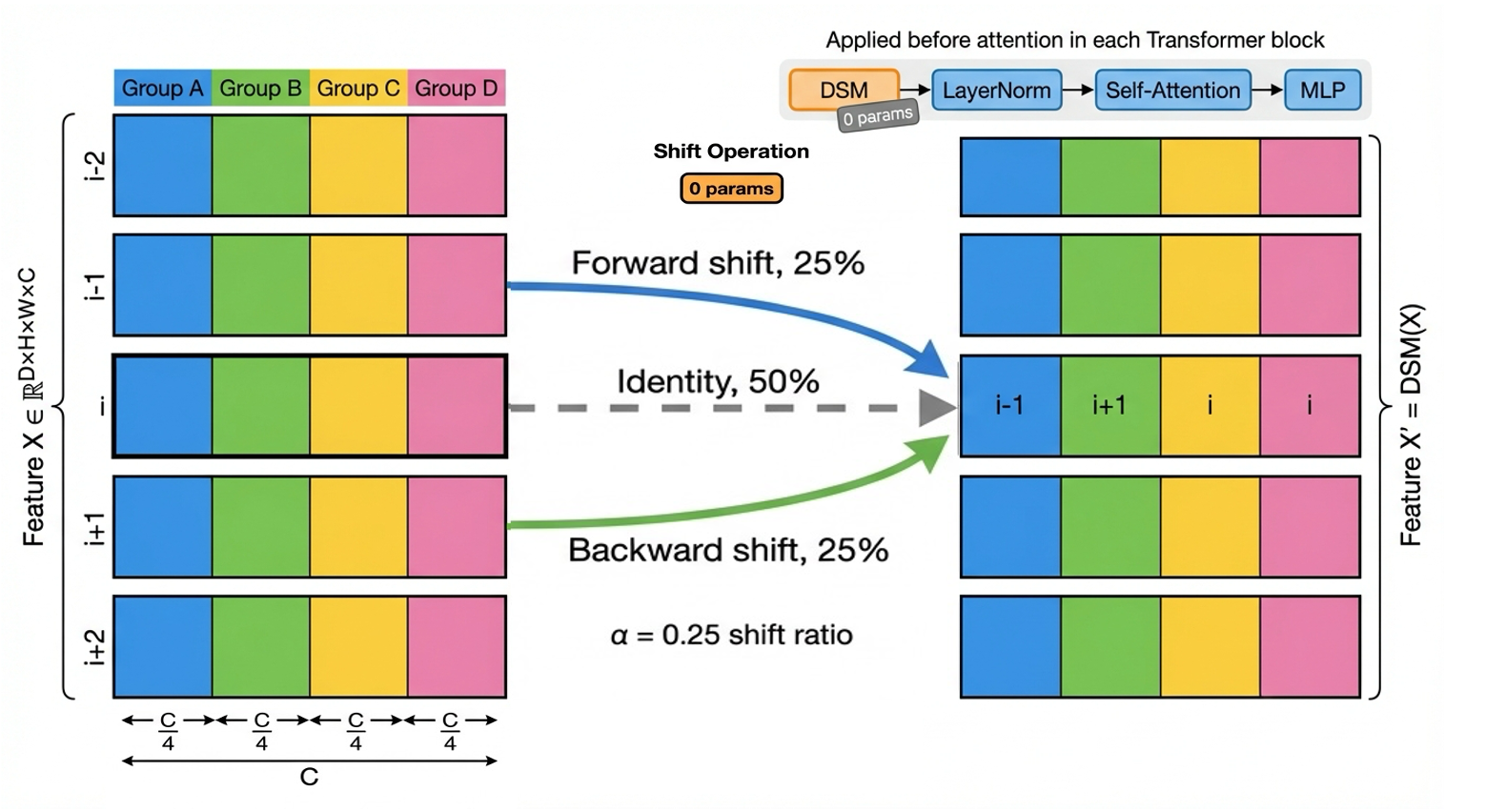}
  \caption{Depth Shift Module (DSM) applied along the depth axis.
  Before self-attention, a fraction $\alpha$ of channels are shifted
  forward from slice $i{-}1$ and backward from slice $i{+}1$, while
  the remaining $(1{-}2\alpha)$ channels stay in place. This gives
  each slice access to neighbouring features at zero extra parameters
  or FLOPs.}
  \label{fig:dsm}
\end{figure}

Processing a volume slice by slice through a 2D encoder isolates each slice from its neighbours, preventing the model from capturing inter-slice continuity. Adding 3D convolutions or cross-slice attention would introduce substantial parameters and compute. We instead introduce the Depth Shift Module (DSM)~\cite{lin2019tsm}---inspired by TSM originally proposed for efficient video understanding---repurposed here for cross-slice feature exchange along the depth axis, as shown in Fig.~\ref{fig:dsm}. DSM adds no learnable parameters and performs no multiply--accumulate (MAC) operations; its only overhead is the memory traffic of a slice-strided copy, which we measure at $0.34$\,ms on an RTX5090 GPU ($1.7\%$ of one encoder block's forward latency). Our contribution is not the operator itself but the lifting strategy: we are the first to show that a temporal-shift operator, applied along the spatial depth axis, both outperforms no-exchange and parameter-richer cross-slice alternatives (supplementary material, Table 8) and transfers unchanged from 2D$\to$3D (SAM) to 2D+T$\to$3D+T (SAM2).

Given a feature tensor $\mathbf{X} \!\in\! \mathbb{R}^{D \times H \times W \times C}$, DSM partitions the channels into three groups controlled by a shift ratio $\alpha$:
\begin{equation}
  \mathbf{X}' = \mathrm{DSM}(\mathbf{X})
  = \mathrm{concat}\!\bigl[
      \mathbf{X}^{\mathrm{fwd}},\;
      \mathbf{X}^{\mathrm{bwd}},\;
      \mathbf{X}^{\mathrm{id}}
    \bigr],
\end{equation}
where $\mathbf{X}^{\mathrm{fwd}}[i]$ copies $\alpha C$ channels from slice $i{-}1$ (forward shift),
$\mathbf{X}^{\mathrm{bwd}}[i]$ copies $\alpha C$ channels from slice $i{+}1$ (backward shift), and the remaining $(1{-}2\alpha)C$ channels are unchanged. Boundary slices retain their own features to avoid artificial discontinuities.

DSM is applied before self-attention in each transformer block, so that query, key, and value projections already incorporate neighbouring-slice information.  A key design choice is that DSM operates exclusively along the depth dimension---across spatial slices within a single volume---rather than across temporal frames.  In the 4D setting, this ensures a clean separation of concerns: depth-local context is handled by DSM, while temporal dynamics are delegated to SAM2's memory mechanism
(\S\ref{sec:3dsam2}).

\subsection{SAM+D for 3D Volumetric Segmentation}
\label{sec:3dsam}

In the 3D setting, we lift SAM's frozen ViT-B from 2D to 3D
(Fig.~\ref{fig:overview}a). Given a 3D volume, we uniformly subsample
$D_s$ depth slices and resize each to match SAM's input resolution.
The slices are processed as a batch through the ViT-B encoder, where
every transformer block is wrapped with: (i)~DSM shift for cross-slice
channel exchange with ratio $\alpha$, (ii)~attention with DRLoRA
corrections $\Delta\mathbf{Q}$, $\Delta\mathbf{V}$ conditioned on
normalised slice position~$z$, followed by (iii)~the standard residual
connection and frozen feedforward network. Layer norms in each block
are left trainable to allow distribution statistics to adapt, while
all attention weights and MLP parameters stay frozen. After encoding,
per-slice features are reassembled into a 3D feature volume through
SAM's neck projection.

For decoding, we study three designs: a Conv3D decoder that directly
upsamples the 3D feature volume, a Multi-Layer Aggregation (MLA)
decoder that fuses features from multiple encoder depths, and SAM's
native prompt-based decoder augmented with LoRA. All three achieve
competitive performance; SAM's prompt-based decoder is our default.
Full decoder specifications are provided in the supplementary material.

When interactive segmentation is desired, we reuse SAM's frozen prompt
encoder and mask decoder. Since SAM has no notion of depth, we
introduce a learnable depth embedding
$\mathbf{z}_{\mathrm{embed}}$ that encodes the relative distance
between each slice and the prompt point along the $z$-axis via a
two-layer MLP:
\begin{equation}
  \mathbf{D}'_d = \mathbf{D}_{\mathrm{prompt}}
    + \mathbf{z}_{\mathrm{embed}}\!\bigl((z_p - d) / D_s\bigr),
  \quad d = 1, \ldots, D_s.
  \label{eq:depth_embed}
\end{equation}
This embedding is added to the dense prompt embeddings before mask
decoding, so the decoder can modulate predictions based on each
slice's proximity to the annotated slice. LoRA is injected into the Q
and V projections of the mask decoder's attention layers. Per-slice
mask logits are then assembled into a 3D volume and refined through a
lightweight 3D convolution head for cross-slice smoothing and
multi-class output.

\subsection{SAM+D for 4D Spatiotemporal Segmentation}
\label{sec:3dsam2}

In the 4D setting, we lift SAM2 from 2D+T video to 3D+T
spatiotemporal segmentation (Fig.~\ref{fig:overview}b). The same
DRLoRA and DSM modules are inserted into SAM2's Hiera
backbone~\cite{ryali2023hiera} in exactly the same way as the 3D
setting---the only architectural difference is that Hiera uses
Q-pooling at stage boundaries to progressively reduce spatial
resolution, so DRLoRA outputs corrections matching the post-projection
dimension at each stage. The depth embedding
(Eq.~\ref{eq:depth_embed}), decoder LoRA injection, and 3D refinement
head are likewise reused unchanged.

The key distinction from the 3D setting is how temporal dynamics are
handled. DRLoRA routes by depth position only, not by time:
$z_{b,t,d} = d/D_s,\;\forall\,b,t$. Temporal consistency is instead
managed through SAM2's memory architecture, which compresses each
frame's predictions into compact tokens, accumulates them in a memory
bank, and cross-attends current-frame features with stored memories
before decoding. The memory encoder and memory bank remain frozen, but
we inject LoRA into the memory attention module so that the
cross-attention between current-frame features and stored memories can
be adapted to the volumetric setting. This is necessary because the
original memory attention was trained for 2D video frames; adapting it
allows the model to learn depth-aware temporal matching patterns
specific to 3D+T data. The improved encoder features from DRLoRA +
DSM, combined with the adapted memory attention, jointly enable more
accurate spatiotemporal propagation.

Training uses a per-frame DiceCE loss without explicit tracking
supervision---temporal consistency emerges from SAM2's memory, as in
its original video segmentation paradigm. The set of trainable
parameters comprises: DRLoRA experts, layer norms, decoder LoRA,
memory attention LoRA, the depth embedding, and the 3D refinement
head. Implementation details are provided in supplementary material.


\section{Experiments}
\label{sec:exp}

\subsection{Experimental Setup}
\label{sec:exp_setup}

\subsubsection{3D Segmentation Datasets.}
We evaluate the 3D setting on four CT tumor segmentation
benchmarks following the protocol of
3DSAM-adapter~\cite{gong20233dsam}.
\textbf{KiTS}~\cite{heller2021state}: 209 / 30 / 61
(train/val/test) abdominal CT scans for kidney tumor segmentation.
\textbf{Pancreas}~\cite{antonelli2022medical}: 196 / 28 / 57
scans from the Medical Segmentation Decathlon for pancreas tumor
segmentation.
\textbf{LiTS}~\cite{bilic2023liver}: 83 / 11 / 24
contrast-enhanced abdominal CT scans for liver tumor segmentation.
\textbf{Colon}~\cite{antonelli2022medical}: 89 / 11 / 26 scans
for colon cancer segmentation.
All datasets use the same train/val/test splits as
3DSAM-adapter~\cite{gong20233dsam} for fair comparison.

\subsubsection{4D Segmentation and Tracking Datasets.}
For the 4D (3D+T) setting we evaluate on
Fluo-N3DH-SIM+~\cite{ulman2017objective} from the Cell Tracking
Challenge (CTC), which contains synthetically generated 3D
time-lapse fluorescence microscopy sequences of HL60 cell nuclei
produced by the CytoPacq/MitoGen simulator~\cite{svoboda2016mitogen}.
Each volume spans approximately $650\times650\times59$ voxels at
29-minute intervals, with cells exhibiting division, migration, and
shape deformation. Ground-truth instance segmentation masks and
lineage tracking annotations are provided.
We train and validate on Sequence~01 and reserve Sequence~02
entirely for testing.
\subsubsection{Evaluation Protocol.}
\label{evl}
For 3D segmentation we report the Dice Similarity Coefficient
(DSC, \%) and the Normalized Surface Distance (NSD, \%) at a
tolerance of 2\,mm, following~\cite{gong20233dsam}.
For 4D segmentation and tracking we adopt the standard CTC
metrics~\cite{ulman2017objective}: DET (detection accuracy),
SEG (segmentation accuracy), TRA (tracking accuracy), and the
overall CTC benchmark score
$\mathrm{OP_{CTB}} = 0.5\,(\mathrm{SEG}+\mathrm{TRA})$.

During 4D inference, cells may divide or newly enter the field of
view. Our tuned SAM2 handles these events by issuing a new point
prompt for each newly divided or appeared cell; otherwise, a single
prompt on the first frame suffices and the model proceeds end-to-end
without post-processing.
We compare under two modes following the tracking-by-detection
paradigm:
\textbf{(1)~GT segmentation + SAM2 linking:} ground-truth masks are available at every frame; for each cell, SAM2's memory produces ranked candidate matches to the next frame, and the known masks are used to validate the link, rejecting matches to non-existent cells and falling back to the next-highest-probability candidate. This isolates linking quality from segmentation.
\textbf{(2)~Cellpose detector + SAM2 propagation:}
Cellpose~\cite{stringer2021cellpose} is used as the off-the-shelf
3D detector to discover new cells and divisions; SAM2 initialises each cell with a point prompt and propagates its mask forward using its memory bank. We use the decoder's native predicted-IoU output as a per-mask confidence estimate to arbitrate between the two sources: high-confidence tracklet masks are trusted, while low-confidence frames fall back to Cellpose's segmentation.

Comparing the two modes quantifies the contribution of SAM2's
learned temporal propagation versus per-frame detection.
Detailed hyperparameter settings are provided in the supplementary
material.

\subsubsection{Implementation Details.}
We use SAM ViT-B as the frozen backbone for 3D and SAM2 Hiera-B+ for
4D, with DRLoRA ($E\!=\!4$ experts, rank $r\!=\!16$) and DSM
($\alpha\!=\!25\%$) inserted into each encoder block.
For the 3D setting, given a point prompt, we crop a
$128 \times 128 \times 128$ volume centered on the prompt location
and replicate it to three channels, yielding an input of size
$3 \times 128 \times 128 \times 128$.
We train for 500 epochs using AdamW with a learning rate of
$4 \times 10^{-4}$ and cosine annealing.
The 3D model has ${\sim}$83.78\,M total parameters, of which only
${\sim}$2.36\,M (${\sim}$2.8\%) are trainable.
For the 4D setting, we additionally inject LoRA (rank $r\!=\!16$)
into all 4 memory attention layers to adapt temporal propagation.
Input volumes are cropped around the target cell with a
$2.0\times$ padding factor, yielding per-timestep inputs of size
$3 \times 32 \times 256 \times 256$, and each training clip spans
$T\!=\!16$ timesteps.
We train for 500 epochs with a learning rate of $1 \times 10^{-4}$.
The 4D model trains ${\sim}$3.11\,M (${\sim}$3.7\%) of
${\sim}$83.90\,M total parameters.
All experiments use 3 NVIDIA RTX 6000 Pro GPUs with mixed-precision
(BF16) training and gradient checkpointing. At inference, SAM+D processes a full
$128^3$ volume in $\sim$228.5\,ms on a single NVIDIA RTX 6000 Pro
GPU. Full hyperparameters are provided in the supplementary
material.

\subsection{3D Volumetric Segmentation Results}
\label{sec:exp_3d}

\begin{figure*}[!htbp]
\centering

\includegraphics[width=0.95\textwidth]{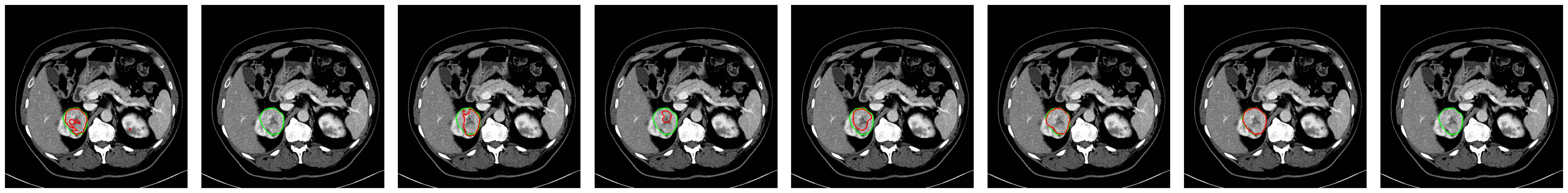}\\
\includegraphics[width=0.95\textwidth]{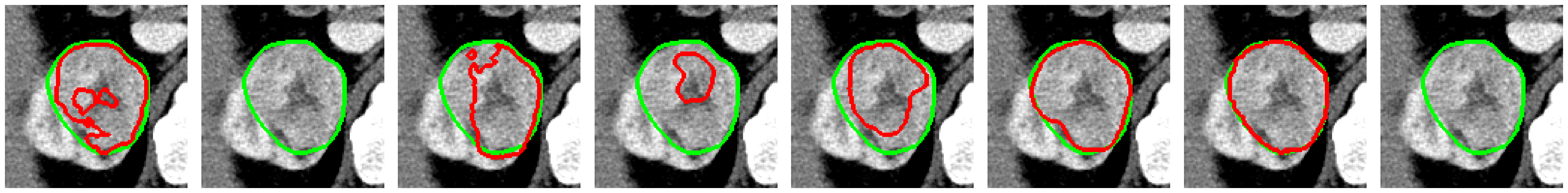}\\

\includegraphics[width=0.95\textwidth]{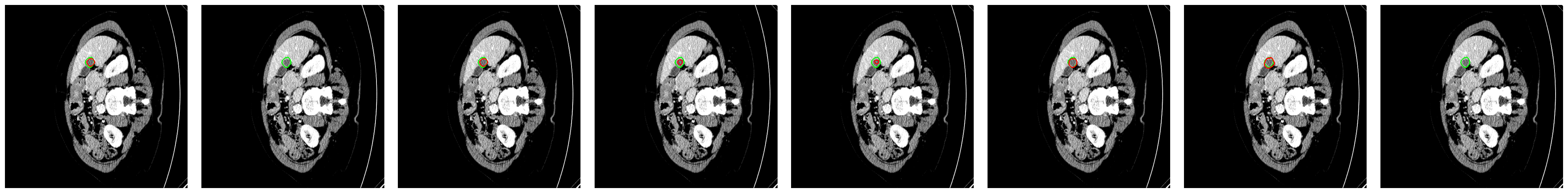}\\
\includegraphics[width=0.95\textwidth]{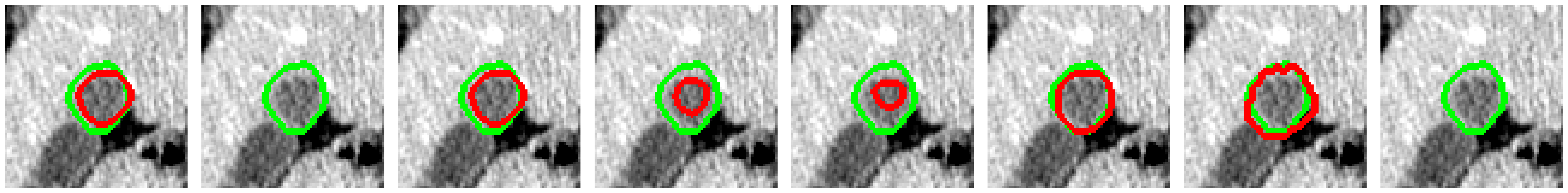}\\

\includegraphics[width=0.95\textwidth]{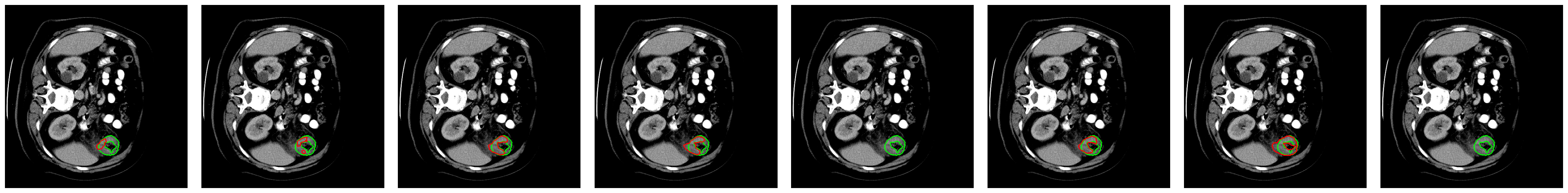}\\
\includegraphics[width=0.95\textwidth]{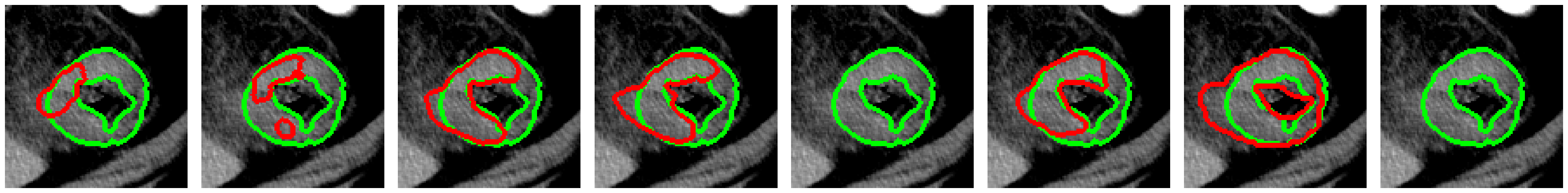}\\

\includegraphics[width=0.95\textwidth]{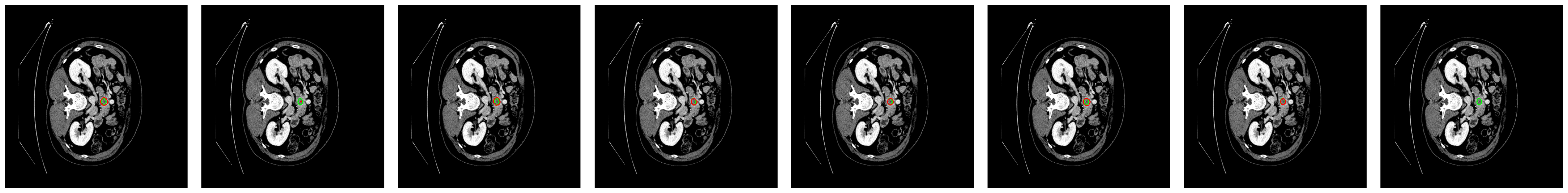}\\
\includegraphics[width=0.95\textwidth]{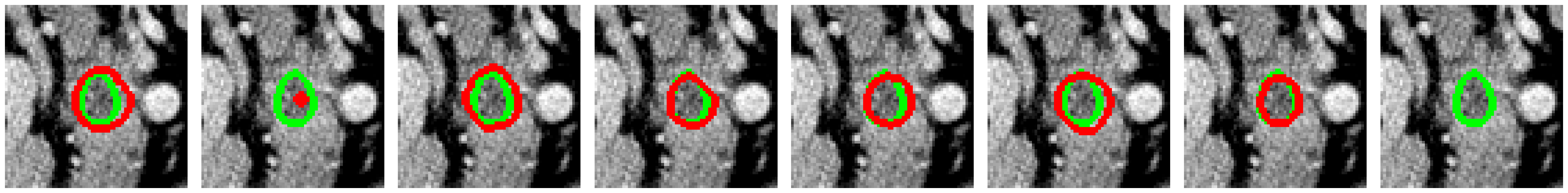}\\[2pt]

\noindent
\begin{tabular}{@{}p{0.119\textwidth}@{}p{0.119\textwidth}@{}p{0.119\textwidth}@{}p{0.119\textwidth}@{}p{0.119\textwidth}@{}p{0.119\textwidth}@{}p{0.119\textwidth}@{}p{0.119\textwidth}@{}}
  \centering\scriptsize ~\cite{lee20223d} &
  \centering\scriptsize ~\cite{zhou2021nnformer} &
  \centering\scriptsize ~\cite{tang2022self} &
  \centering\scriptsize ~\cite{wang2021transbts} &
  \centering\scriptsize ~\cite{shaker2024unetrpp} &
  \centering\scriptsize ~\cite{gong20233dsam} &
  \centering\scriptsize \textbf{Ours} &
  \centering\arraybackslash\scriptsize GT
\end{tabular}
\noindent
\caption{%
  \textbf{Qualitative 3D segmentation results.}
  Each pair shows a full slice (top) and an enlarged crop
  (bottom) for four representative cases.
  \textcolor{green}{Green}: GT;
  \textcolor{red}{Red}: prediction.
}
\label{fig:qual_3d}
\end{figure*}

Table~\ref{tab:3d_main} compares SAM+D with all baselines on the four CT tumor segmentation benchmarks.
All SAM+D results use a single point prompt per volume. We compare against three categories of methods
following the evaluation protocol of 3DSAM-adapter~\cite{gong20233dsam}: (i)~Fully-supervised 3D methods: nnU-Net~\cite{isensee2021nnu}, TransBTS~\cite{wang2021transbts}, nnFormer~\cite{zhou2021nnformer}, Swin-UNETR~\cite{tang2022self}, UNETR++~\cite{shaker2024unetrpp}, and 3D UX-Net~\cite{lee20223d}; (ii)~Interactive point-prompt methods: Gaussian Kernel~\cite{xu2016deep},
MIDeepSeg~\cite{luo2021mideepseg}, GPCIS~\cite{zhou2023interactive}, and Visual Sampler~\cite{zou2023segment}; (iii)~SAM-based methods: SAM-B~\cite{kirillov2023segment} applied per-slice without volumetric adaptation, 3DSAM-adapter~\cite{gong20233dsam}, and the volumetric SAM adapters Med-SA~\cite{wu2025medical} and MA-SAM~\cite{chen2024ma}. Baseline numbers for categories (i), (ii), and SAM-B/3DSAM-adapter are taken directly
from~\cite{gong20233dsam}; Med-SA and MA-SAM are fine-tuned per dataset under the same splits.
As shown in the table, SAM+D achieves near state-of-the-art performance across virtually every evaluation metric on all four datasets, consistently matching or surpassing both fully-supervised 3D methods and existing SAM-based approaches despite using only a single point prompt. Among SAM-based methods, SAM+D surpasses the directly comparable Med-SA and MA-SAM on all four datasets; we omit SAM-Med3D~\cite{wang2025sam}, SegVol~\cite{du2024segvol}, and MedSAM2~\cite{ma2025medsam2} here because our test splits fall within their pretraining corpora (zero-shot numbers and a contamination-free private liver-cancer CT evaluation are in the supplementary material).
We evaluate the three decoder variants described in (\S\ref{sec:3dsam}): a Conv3D decoder, a multi-layer aggregation (MLA) decoder, and the SAM prompt decoder with LoRA. Among the three variants, the SAM prompt decoder achieves the highest average Dice and NSD across all four datasets while
requiring the fewest trainable parameters (2.57\,M), and is therefore used as our default. Please refer to supplementary material for detail.
We show qualitative comparisons on one representative case from each of the four datasets in Figure~\ref{fig:qual_3d}.

\begin{table}[!t]
\centering
\caption{%
   Comparison with fully-supervised, interactive, and SAM-based methods on four tumor segmentation datasets. Dice (\%) and NSD (\%) are reported
}
\label{tab:3d_main}
\resizebox{\textwidth}{!}{%
\begin{tabular}{l c c c c c c c c c}
\toprule
\multirow{2}{*}{Methods}
  & \multicolumn{2}{c}{Kidney Tumor}
  & \multicolumn{2}{c}{Pancreas Tumor}
  & \multicolumn{2}{c}{Liver Tumor}
  & \multicolumn{2}{c}{Colon Cancer}
  & \multirow{2}{*}{\#Trainable Params} \\
\cmidrule(lr){2-3} \cmidrule(lr){4-5} \cmidrule(lr){6-7} \cmidrule(lr){8-9}
  & Dice$\uparrow$ & NSD$\uparrow$ & Dice$\uparrow$ & NSD$\uparrow$ & Dice$\uparrow$ & NSD$\uparrow$ & Dice$\uparrow$ & NSD$\uparrow$ & \\
\midrule
nnU-Net~\cite{isensee2021nnu}
  & 73.09 & 77.42
  & 41.70 & 62.92
  & 60.27 & \underline{75.60}
  & 44.38 & 53.66
  & 30.76\,M \\
TransBTS~\cite{wang2021transbts}
  & 41.70 & 38.37
  & 32.00 & 46.91
  & 35.10 & 50.08
  & 17.62 & 22.03
  & 32.33\,M \\
nnFormer~\cite{zhou2021nnformer}
  & 44.78 & 43.17
  & 35.40 & 53.30
  & 35.04 & 44.53
  & 22.04 & 30.72
  & 149.49\,M \\
Swin-UNETR~\cite{tang2022self}
  & 66.48 & 72.44
  & 39.66 & 58.33
  & 52.16 & 65.71
  & 33.53 & 41.16
  & 62.19\,M \\
UNETR++~\cite{shaker2024unetrpp}
  & 57.03 & 60.09
  & 37.59 & 53.33
  & 38.74 & 52.76
  & 25.53 & 30.41
  & 55.70\,M \\
3D UX-Net~\cite{lee20223d}
  & 58.98 & 59.60
  & 34.92 & 52.39
  & 47.41 & 62.90
  & 28.62 & 35.20
  & 53.01\,M \\
\midrule
Gaussian Kernel (1\,pt/vol)~\cite{xu2016deep}
  & 22.32 & 37.27
  & 36.60 & 56.79
  & 39.27 & 53.84
  & 22.54 & 36.41
  & 31.20\,M \\
Gaussian Kernel (3\,pt/vol)~\cite{xu2016deep}
  & 70.56 & 75.77
  & 53.90 & 73.34
  & 54.63 & 71.48
  & 58.99 & 74.22
  & 31.20\,M \\
MIDeepSeg (1\,pt/vol)~\cite{luo2021mideepseg}
  & 41.65 & 46.98
  & 39.10 & 64.49
  & 31.99 & 43.08
  & 45.17 & 60.67
  & 31.20\,M \\
MIDeepSeg (3\,pt/vol)~\cite{luo2021mideepseg}
  & 60.54 & 64.48
  & 48.58 & 77.09
  & 39.58 & 56.58
  & 59.36 & 76.45
  & 31.20\,M \\
GPCIS (1\,pt/vol)~\cite{zhou2023interactive}
  & 61.54 & 71.03
  & 41.18 & 66.12
  & 41.21 & 58.11
  & 53.11 & 72.45
  & 31.29\,M \\
GPCIS (3\,pt/vol)~\cite{zhou2023interactive}
  & 68.15 & 79.68
  & 48.33 & 76.99
  & 48.40 & 70.55
  & 57.74 & 77.16
  & 31.29\,M \\
Visual Sampler (1\,pt/vol)~\cite{zou2023segment}
  & 72.47 & 78.25
  & 49.48 & 69.49
  & 52.51 & 64.29
  & 53.49 & 68.05
  & 33.67\,M \\
Visual Sampler (3\,pt/vol)~\cite{zou2023segment}
  & 73.77 & 79.87
  & 51.13 & 72.67
  & 54.85 & 66.73
  & 57.52 & 71.64
  & 33.67\,M \\
\midrule
SAM-B (1\,pt/slice)~\cite{kirillov2023segment}
  & 36.30 & 29.86
  & 24.01 & 26.74
  & 6.71  & 7.63
  & 28.83 & 33.63
  & -- \\
3DSAM-adapter (1\,pt/vol)~\cite{gong20233dsam}
  & 80.16 & 87.40
  & 53.69 & 76.44
  & 58.02 & 71.55
  & 59.29 & 76.10
  & 25.46\,M \\
SAM-B (3\,pts/slice)~\cite{kirillov2023segment}
  & 39.66 & 34.85
  & 29.80 & 33.24
  & 7.87  & 6.76
  & 35.26 & 39.31
  & -- \\
3DSAM-adapter (3\,pts/vol)~\cite{gong20233dsam}
  & 81.50 & 88.18
  & \underline{54.82} & \underline{78.60}
  & \underline{61.25} & \textbf{77.09}
  & 60.93 & 77.56
  & 25.46\,M \\
Med-SA (1\,pt/slice)~\cite{wu2025medical}
  & 70.73 & 82.41 
  & 43.59 & 73.35 
  & 51.61 & 72.04 
  & 53.56 & 74.85 
  & 13.00\,M \\
MA-SAM (no prompt)~\cite{chen2024ma}
  & 69.25 & 65.84 
  & 34.72 & 47.72 
  & 56.28 & 62.87 
  & 50.32 & 56.37 
  & 63.04\,M \\
\midrule
\textbf{Ours -- Conv3D dec.\ (1\,pt/vol)}
  & 82.42 & 88.89
  & 50.02 & 67.92
  & 54.06 & 62.02
  & \textbf{66.39} & \textbf{79.85}
  & \underline{3.29\,M} \\
\textbf{Ours -- MLA dec.\ (1\,pt/vol)}
  & \underline{84.16} & \underline{91.07}
  & 51.34 & 70.52
  & 58.54 & 67.68
  & \underline{63.70} & \underline{78.42}
  & 10.27\,M \\
\textbf{Ours -- SAM dec.\ (1\,pt/vol)}
  & \textbf{84.74} & \textbf{92.11}
  & \textbf{59.68} & \textbf{79.05}
  & \textbf{63.33} & 73.52
  & \underline{63.70} & 77.69
  & \textbf{2.57\,M} \\
\bottomrule
\end{tabular}%
}
\end{table}

\begin{table}[!t]
\centering
\caption{4D segmentation and tracking results on Fluo-N3DH-SIM+
from the Cell Tracking Challenge. The CTC provides two annotated
sequences; we train and validate on Seq\,01 and test on Seq\,02.
SEG (\%), DET (\%), TRA (\%) and $\mathrm{OP_{CTB}}$ (\%) are
reported. '--' in \#Trainable Params indicates training-free
methods.}
\label{tab:4d_main}
\resizebox{\columnwidth}{!}{%
\begin{tabular}{l l c c c c c}
\toprule
Method & Seg.\ Source & SEG$\uparrow$ & DET$\uparrow$ & TRA$\uparrow$ & $\mathrm{OP_{CTB}}$$\uparrow$ & \#Trainable Params \\
\midrule
\multirow{2}{*}{BGU-IL~\cite{ben2022graph}}
  & GT       & --    & --    & 99.96 & 99.98 & \multirow{2}{*}{127.70\,M} \\
  & CellPose & 53.80 & 85.20 & 84.20 & 69.00 & \\
\midrule
\multirow{2}{*}{Ultrack~\cite{bragantini2024ucmtracking, bragantini2024ultrack}}
  & GT       & --    & --    & 99.23 & 99.36 & \multirow{2}{*}{--} \\
  & CellPose & 54.51 & 85.43 & 85.17 & 69.84 & \\
\midrule
\multirow{2}{*}{TrackStra~\cite{gallusser2024trackastra}}
  & GT       & --    & --    & \textbf{99.98} & \textbf{99.99} & \multirow{2}{*}{6.90\,M} \\
  & CellPose & 54.15 & 84.86 & 84.16 & 69.16 & \\
\midrule
\multirow{2}{*}{SAM2 (slice-based)~\cite{chen2025segment}}
  & GT       & --    & --    & 98.18 & 98.20 & \multirow{2}{*}{--} \\
  & CellPose & 54.19 & 84.84 & 84.14 & 69.16 & \\
\midrule
\multirow{2}{*}{\textbf{SAM2+D (Ours)}}
  & GT       & --    & --    & 99.31 & 99.47 & \multirow{2}{*}{\textbf{3.11\,M}} \\
  & CellPose & \textbf{56.79} & \textbf{87.32} & \textbf{87.15} & \textbf{71.97} & \\
\bottomrule
\end{tabular}%
}
\end{table}

\definecolor{vivpurple}{RGB}{180,0,255}

\begin{figure*}[!htbp]
\centering

\includegraphics[width=0.15\textwidth]{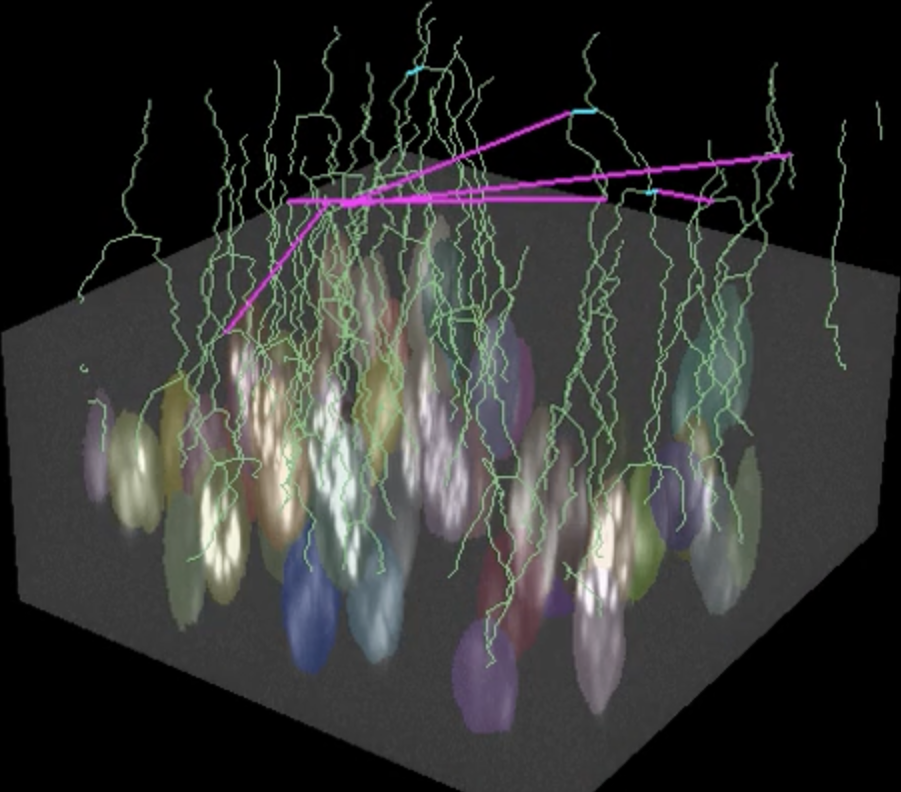}%
\includegraphics[width=0.15\textwidth]{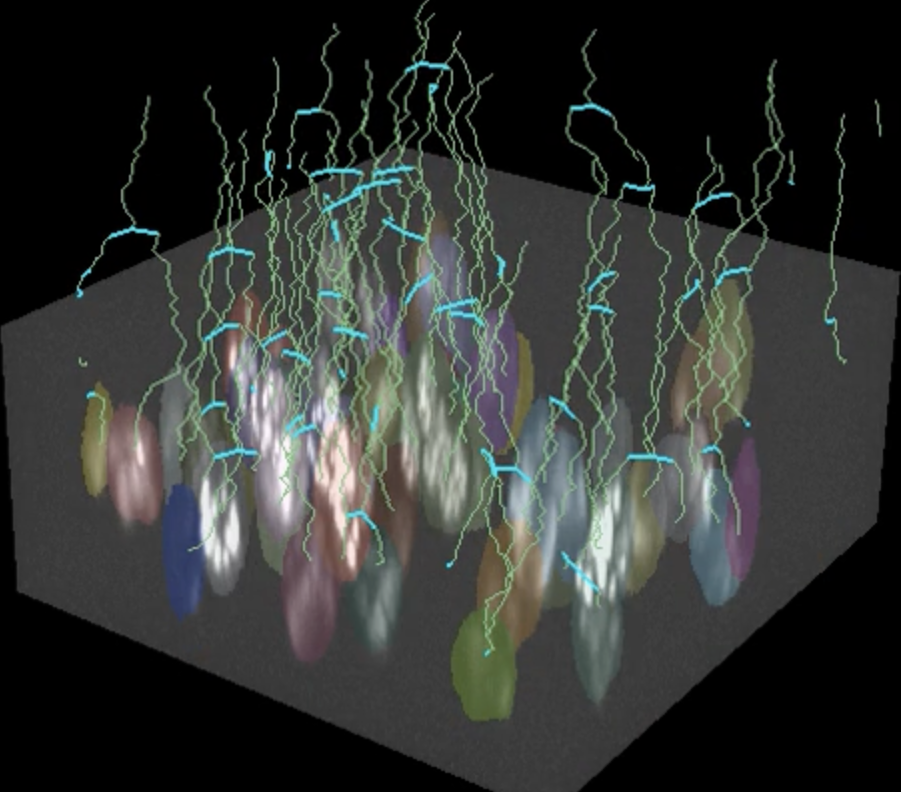}%
\includegraphics[width=0.15\textwidth]{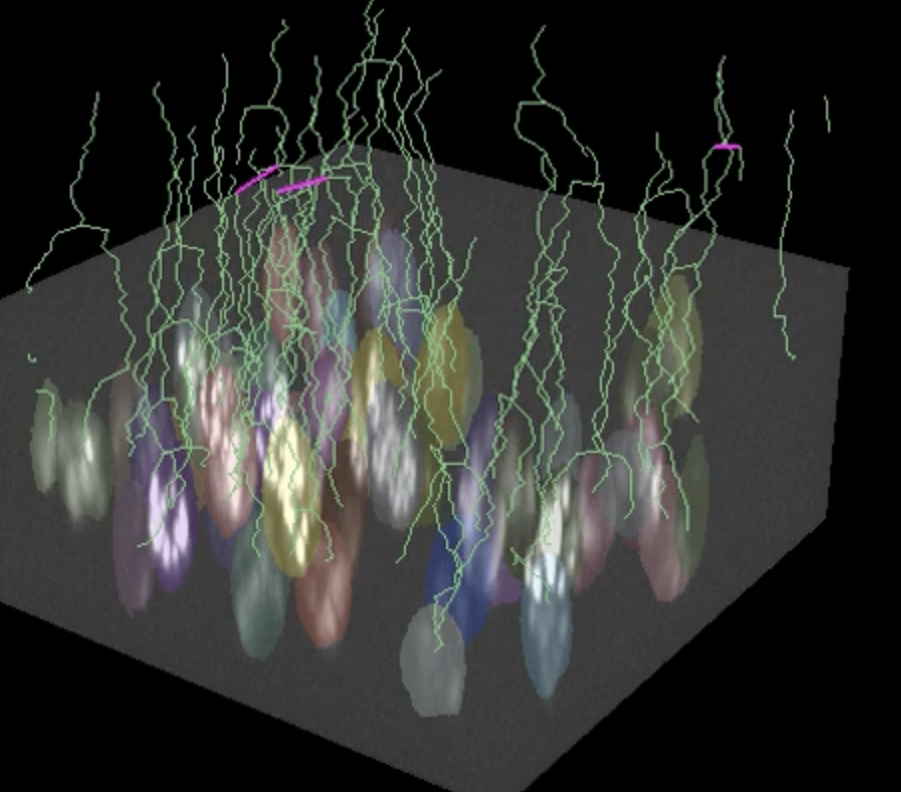}%
\includegraphics[width=0.15\textwidth]{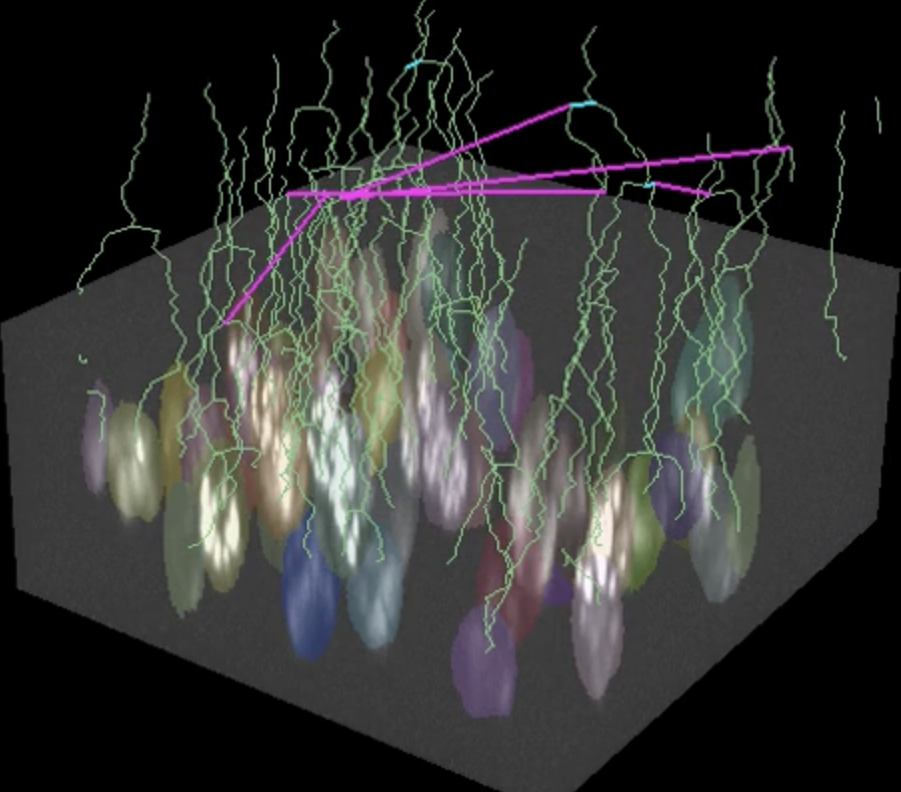}%
\includegraphics[width=0.15\textwidth]{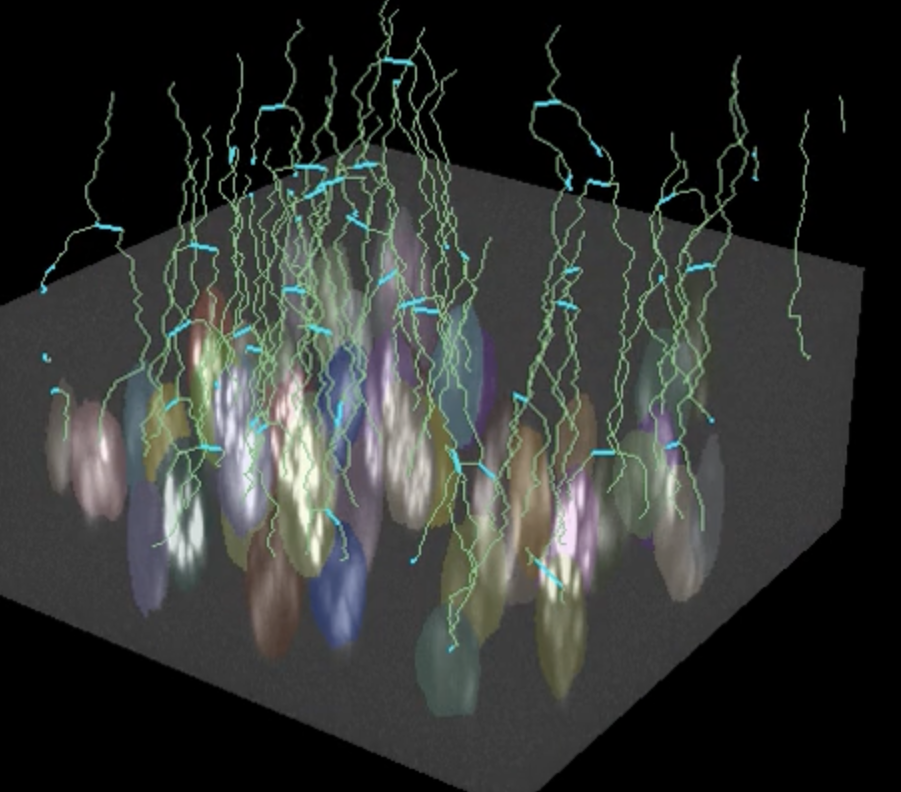}%
\includegraphics[width=0.15\textwidth]{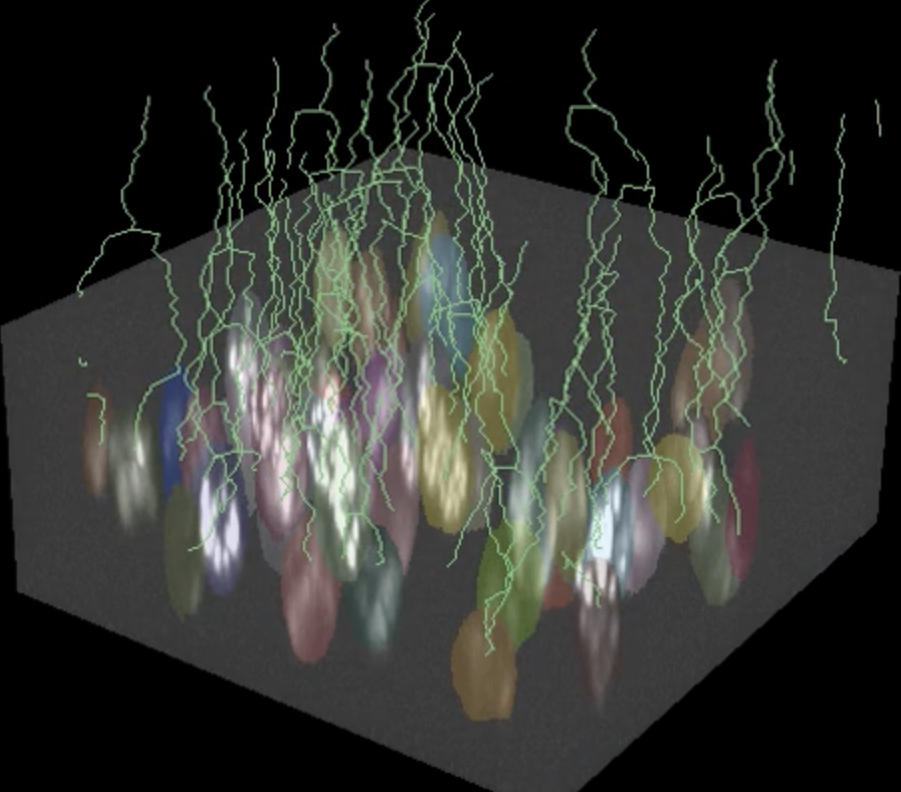}\\[2pt]

\includegraphics[width=0.15\textwidth]{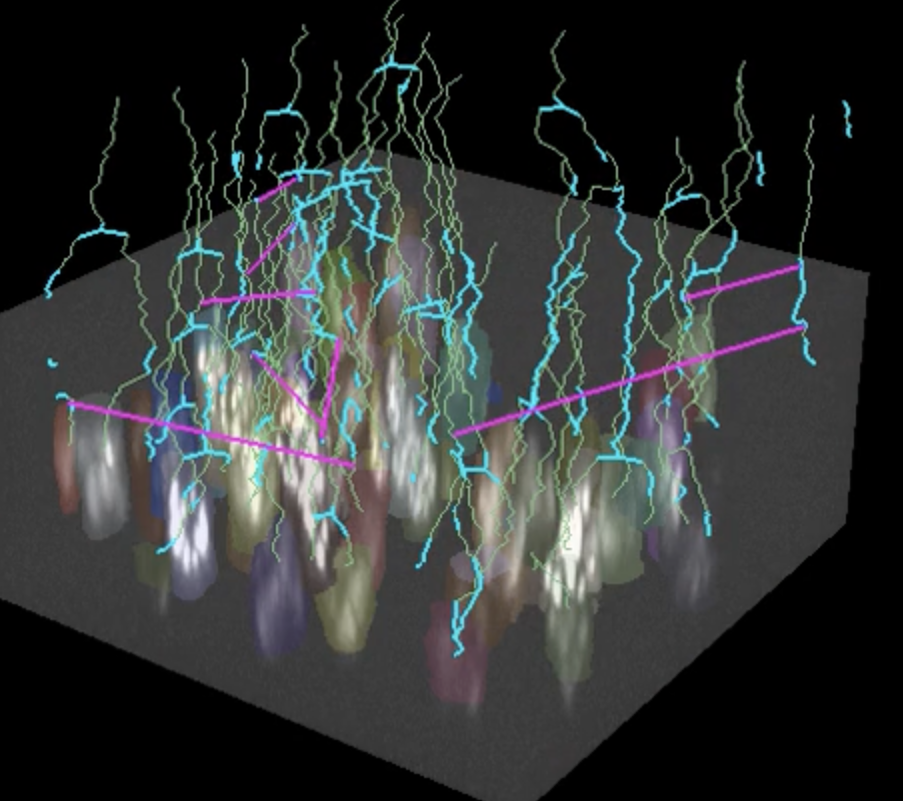}%
\includegraphics[width=0.15\textwidth]{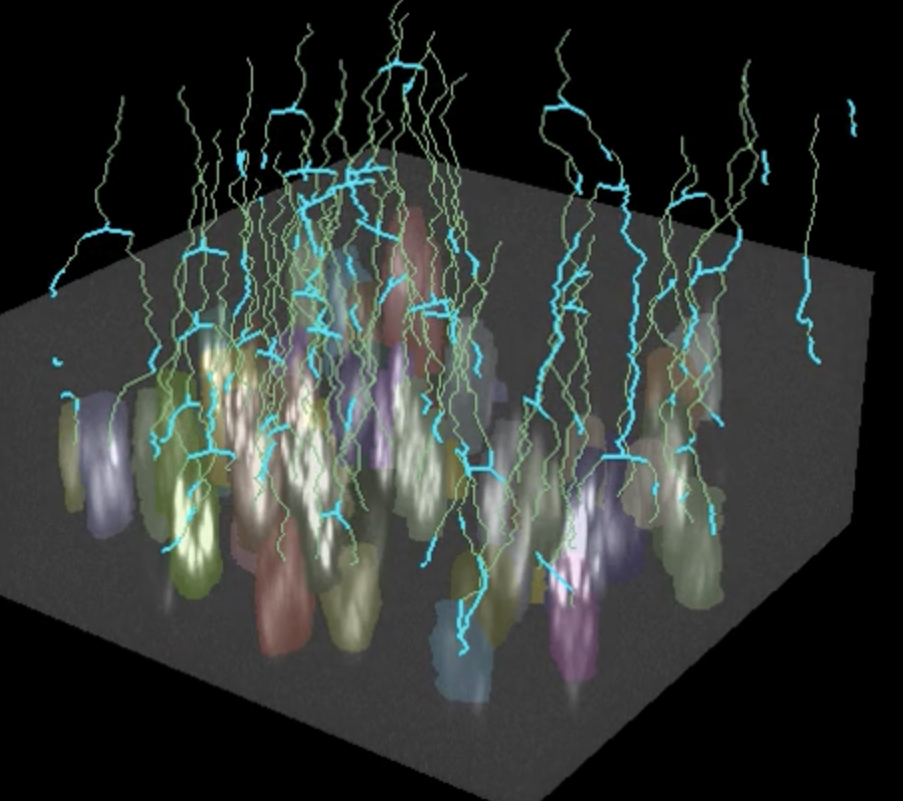}%
\includegraphics[width=0.15\textwidth]{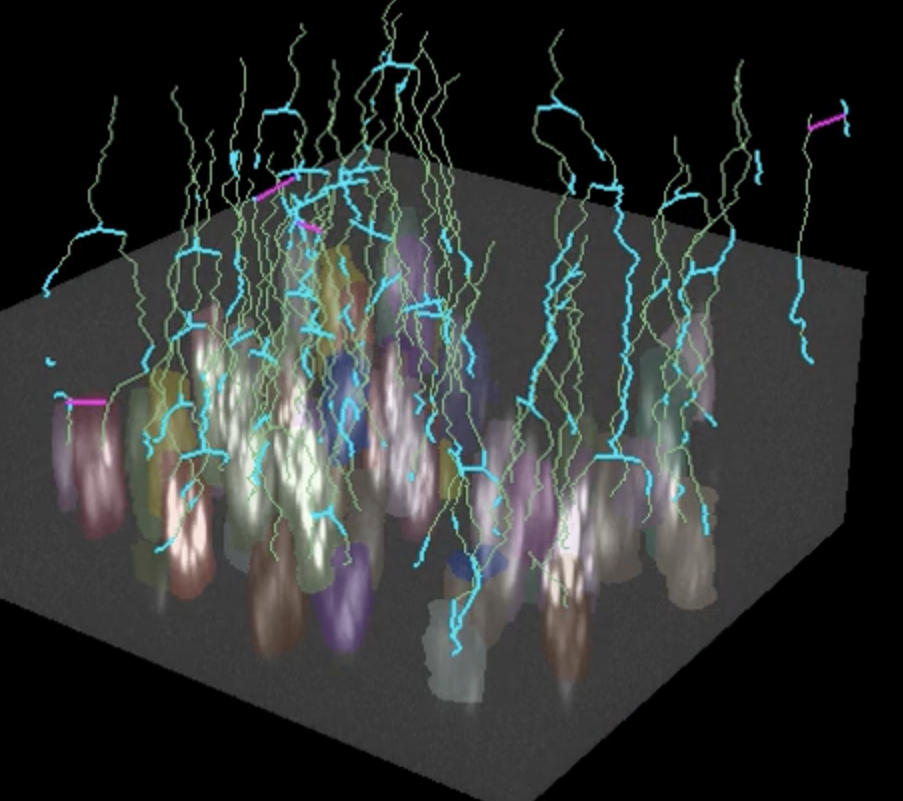}%
\includegraphics[width=0.15\textwidth]{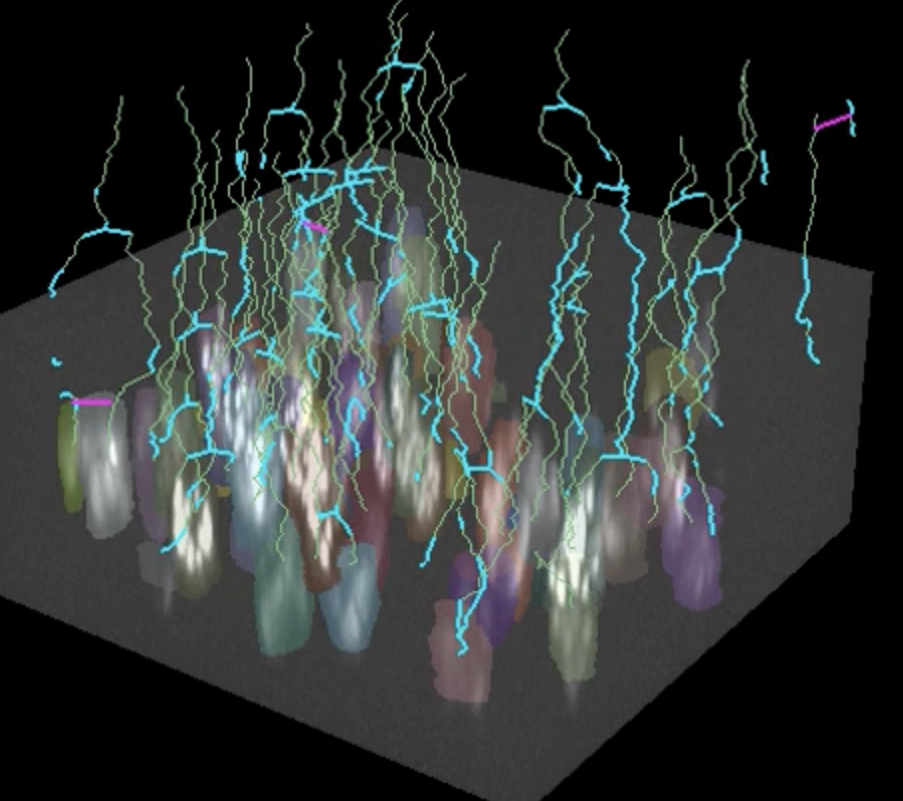}%
\includegraphics[width=0.15\textwidth]{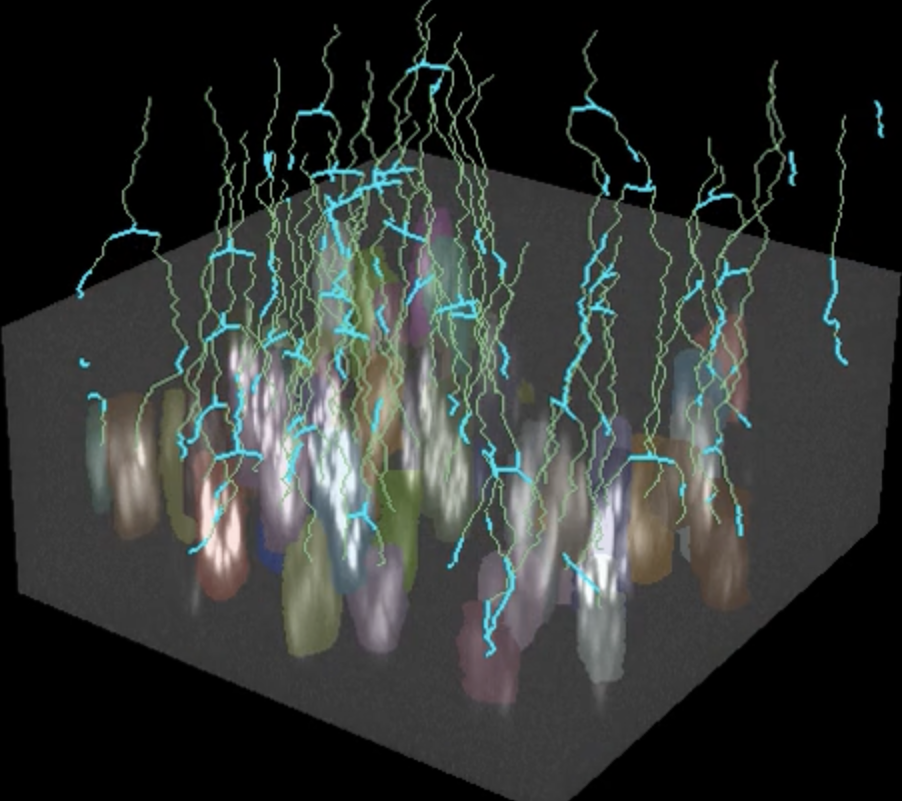}%
\includegraphics[width=0.15\textwidth]{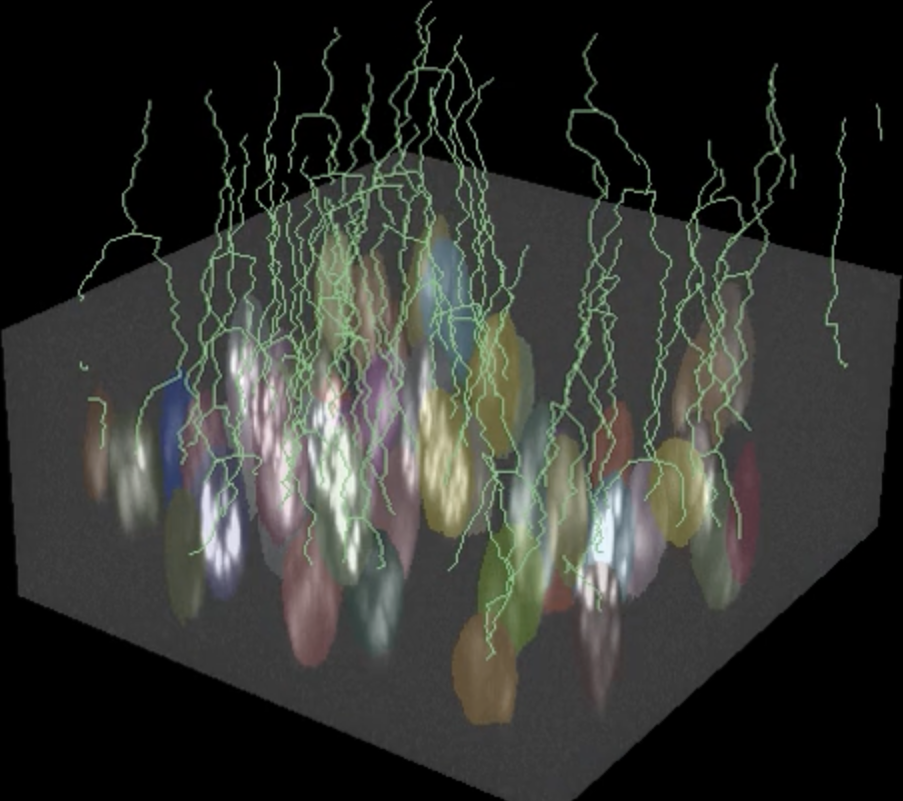}\\[2pt]

\noindent
\begin{tabular}{@{}p{0.15\textwidth}@{}p{0.15\textwidth}@{}p{0.15\textwidth}@{}p{0.15\textwidth}@{}p{0.15\textwidth}@{}p{0.15\textwidth}@{}}
  \centering\scriptsize BGU-IL~\cite{ben2022graph} &
  \centering\scriptsize Ultrack~\cite{bragantini2024ultrack} &
  \centering\scriptsize TrackStra~\cite{gallusser2024trackastra} &
  \centering\scriptsize SAM2~\cite{chen2025segment} &
  \centering\scriptsize \textbf{Ours} &
  \centering\arraybackslash\scriptsize GT
\end{tabular}
\noindent
\caption{%
  \textbf{Visualization on Fluo-N3DH-SIM+.}
  Top: Mode\,1 (GT segmentation + propagation).
  Bottom: Mode\,2 (Cellpose~\cite{stringer2021cellpose} + propagation).
  \textcolor{green}{Green}: correctly tracked;
  \textcolor{vivpurple}{Purple}: false positive;
  \textcolor{cyan}{Blue}: false negative.
  Best viewed when zoomed in.
}
\label{fig:qual_4d}
\end{figure*}

\subsection{4D Spatiotemporal Segmentation and Tracking Results}
\label{sec:exp_4d}

Table~\ref{tab:4d_main} presents the 4D (3D+T) segmentation and
tracking results on Fluo-N3DH-SIM+. As shown in
Figure~\ref{fig:qual_4d}, our method yields more coherent
segmentation than the slice-based SAM2 approach~\cite{chen2025segment}
without requiring dedicated tracking algorithms.
Notably, under the Cellpose setting all baseline methods produce nearly identical SEG scores ($\sim$54\%), since their segmentation quality is fully determined by the shared Cellpose detections. SAM2+D breaks
this ceiling (56.79\%), demonstrating that temporal propagation
can refine per-frame segmentation beyond the detector's quality.
The identical DRLoRA and DSM modules improve SAM2 on 3D time-lapse
tracking just as they improve SAM on medical CT volumes, validating
that dimensional lifting is a general method.


\begin{table}[!htbp]
\centering
\caption{Ablation study on the LiTS dataset. All models use
SAM (ViT-B) and SAM's original decoder. Default settings are marked with *.}
\label{tab:ablation}
\begin{minipage}[t]{0.48\textwidth}
\centering
\begin{tabular}{lcc}
\toprule
\textbf{Setting} & \textbf{Dice(\%)} & \textbf{NSD(\%)} \\
\midrule
\multicolumn{3}{l}{\textit{(a) Component contribution}} \\
DRLoRA only              & 56.41 & 65.28 \\
\,+ DSM                  & 59.17 & 68.74 \\
\,+ z-embed              & 61.05 & 71.03 \\
\,+ Dec.\ LoRA *   & 63.33 & 73.52 \\
\midrule
\multicolumn{3}{l}{\textit{(b) DSM shift ratio $\alpha$}} \\
$\alpha=0.125$           & 61.22 & 71.46 \\
$\alpha=0.25$*           & 63.33 & 73.52 \\
$\alpha=0.5$             & 62.08 & 72.15 \\
\bottomrule
\end{tabular}
\end{minipage}%
\hfill
\begin{minipage}[t]{0.48\textwidth}
\centering
\begin{tabular}{lcc}
\toprule
\textbf{Setting} &  \textbf{Dice(\%)} & \textbf{NSD(\%)} \\
\midrule
\multicolumn{3}{l}{\textit{(c) DRLoRA experts $K$}} \\
$K=1$ (std.\ LoRA)      & 58.86 & 68.21 \\
$K=2$                    & 61.47 & 71.30 \\
$K=4$*                   & 63.33 & 73.52 \\
$K=8$                    & 63.10 & 73.18 \\
\midrule
\multicolumn{3}{l}{\textit{(d) LoRA rank $r$}} \\
$r=4$                    & 60.72 & 70.43 \\
$r=8$                   & 62.85 & 73.01 \\
$r=16$*                   & 63.33 & 73.52 \\
$r=32$                   & 62.19 & 72.34 \\

\bottomrule
\end{tabular}
\end{minipage}
\end{table}

\subsection{Ablation Studies}
\label{sec:exp_ablation}

We ablate each component on the LiTS benchmark (1\,pt/volume) in
Table~\ref{tab:ablation}. The component study~(a) shows that each
module contributes complementary gains: DRLoRA adapts \textbf{what}
features are extracted at each depth, DSM controls \textbf{how} they
exchange information across slices, and the decoder components
refine mask generation with depth-aware cues. The default
hyperparameters ($\alpha\!=\!0.25$, $K\!=\!4$, $r\!=\!16$)
consistently yield the best performance across all sweeps~(b--d). Depth routing also outperforms content-based MoE-LoRA routers (MoLoRA, MoLE,
MixLoRA) by $+7$--$9$ Dice on LiTS under a matched budget (supplementary material, Table 10).

\section{Discussion}
\label{sec:discussion}

We have not yet evaluated SAM+D on SAM3~\cite{carion2025sam}, which
introduces text-prompted multimodal segmentation; directly applying
our modules may not suffice for cross-modal adaptation, and we leave
this to future work. Additionally, our 3D pipeline treats a single
anatomical axis as depth; averaging predictions across all three
axes (sagittal, coronal, transverse) could further improve quality.

For the 4D setting, our method runs end-to-end from a single prompt
when tracking a single object. In practice, however, frequent cell
divisions and new entries require per-event prompting, increasing
inference cost and making fully automatic tracking difficult without
an external detector. Moreover, 4D training is computationally demanding ($\sim$876 GPU-hours per run), which constrained the breadth of our 4D experiments.

In this work, we lift a 2D model to 3D and a 2D+T model to 3D+T
by injecting lightweight depth-aware modules along a new axis. The
same principle should naturally extend to the temporal dimension:
by treating time as the added axis, a pre-trained 2D image model
could be lifted to handle video without architecture-level changes.
We believe this paradigm of adding one dimension at a time through
parameter-efficient adaptation offers a practical path toward
extending foundation models to higher-dimensional tasks.

\section{Conclusion}
\label{sec:conclusion}

We presented SAM+D, a parameter-efficient framework that lifts
pre-trained 2D foundation models to 3D and 2D+T models to 3D+T
through two lightweight modules: Depth-Routed LoRA (DRLoRA), which provides depth-conditioned feature adaptation by routing low-rank experts on slice depth, and Depth Shift Module (DSM), which enables zero-parameter inter-slice feature exchange. With fewer than 4\% trainable parameters, SAM+D achieves competitive performance against
existing methods on four 3D medical segmentation benchmarks
and demonstrates effective spatiotemporal tracking on the Cell
Tracking Challenge. We hope SAM+D provides a practical and
generalisable recipe for extending pre-trained models to
higher-dimensional tasks.

\section*{Acknowledgements}
This work is supported in part by JST CREST, Japan, under Grant JPMJCR25T4, and in part by JSPS KAKENHI, Japan, under Grant 26H00455. We would like to thank Prof. Keiji Nakajima from the Nara Institute of Science and Technology for the valuable support throughout this research.

%
%
\bibliographystyle{splncs04}

\bibliography{main}

@String(ICLR  = {Int. Conf. Learn. Represent.})

@String(ICLR  = {ICLR})

@article{kingma2013auto,
  title={Auto-encoding variational bayes},
  author={Kingma, Diederik P and Welling, Max},
  journal={arXiv preprint arXiv:1312.6114},
  year={2013}
}

@article{oquab2023dinov2,
  title={Dinov2: Learning robust visual features without supervision},
  author={Oquab, Maxime and Darcet, Timoth{\'e}e and Moutakanni, Th{\'e}o and Vo, Huy and Szafraniec, Marc and Khalidov, Vasil and Fernandez, Pierre and Haziza, Daniel and Massa, Francisco and El-Nouby, Alaaeldin and others},
  journal={arXiv preprint arXiv:2304.07193},
  year={2023}
}

@inproceedings{rombach2022high,
  title={High-resolution image synthesis with latent diffusion models},
  author={Rombach, Robin and Blattmann, Andreas and Lorenz, Dominik and Esser, Patrick and Ommer, Bj{\"o}rn},
  booktitle={IEEE Conf. Comput. Vis. Pattern Recog.},
  pages={10684--10695},
  year={2022}
}

@inproceedings{kirillov2023segment,
  title={Segment anything},
  author={Kirillov, Alexander and Mintun, Eric and Ravi, Nikhila and Mao, Hanzi and Rolland, Chloe and Gustafson, Laura and Xiao, Tete and Whitehead, Spencer and Berg, Alexander C and Lo, Wan-Yen and others},
  booktitle={IEEE Conf. Comput. Vis. Pattern Recog.},
  pages={4015--4026},
  year={2023}
}

@article{ravi2024sam,
  title={Sam 2: Segment anything in images and videos},
  author={Ravi, Nikhila and Gabeur, Valentin and Hu, Yuan-Ting and Hu, Ronghang and Ryali, Chaitanya and Ma, Tengyu and Khedr, Haitham and R{\"a}dle, Roman and Rolland, Chloe and Gustafson, Laura and others},
  journal={arXiv preprint arXiv:2408.00714},
  year={2024}
}

@inproceedings{cciccek20163d,
  title={3D U-Net: learning dense volumetric segmentation from sparse annotation},
  author={{\c{C}}i{\c{c}}ek, {\"O}zg{\"u}n and Abdulkadir, Ahmed and Lienkamp, Soeren S and Brox, Thomas and Ronneberger, Olaf},
  booktitle={International conference on medical image computing and computer-assisted intervention},
  pages={424--432},
  year={2016},
  organization={Springer}
}

@article{isensee2021nnu,
  title={nnU-Net: A Self-Configuring Method for Deep Learning-Based Biomedical Image Segmentation},
  author={Isensee, Fabian and Jaeger, Paul F and Kohl, Simon A A and Petersen, Jens and Maier-Hein, Klaus H},
  journal={Nature Methods},
  volume={18},
  number={2},
  pages={203--211},
  year={2021}
}

@article{wang2025sam,
  title={SAM-Med3D: a vision foundation model for general-purpose segmentation on volumetric medical images},
  author={Wang, Haoyu and Guo, Sizheng and Ye, Jin and Deng, Zhongying and Cheng, Junlong and Li, Tianbin and Chen, Jianpin and Su, Yanzhou and Huang, Ziyan and Shen, Yiqing and others},
  journal={IEEE Transactions on Neural Networks and Learning Systems},
  year={2025},
  publisher={IEEE}
}

@article{du2024segvol,
  title={Segvol: Universal and interactive volumetric medical image segmentation},
  author={Du, Yuxin and Bai, Fan and Huang, Tiejun and Zhao, Bo},
  journal={Advances in Neural Information Processing Systems},
  volume={37},
  pages={110746--110783},
  year={2024}
}

@article{ma2025medsam2,
  title={Medsam2: Segment anything in 3d medical images and videos},
  author={Ma, Jun and Yang, Zongxin and Kim, Sumin and Chen, Bihui and Baharoon, Mohammed and Fallahpour, Adibvafa and Asakereh, Reza and Lyu, Hongwei and Wang, Bo},
  journal={arXiv preprint arXiv:2504.03600},
  year={2025}
}

@article{chen2024ma,
  title={Ma-sam: Modality-agnostic sam adaptation for 3d medical image segmentation},
  author={Chen, Cheng and Miao, Juzheng and Wu, Dufan and Zhong, Aoxiao and Yan, Zhiling and Kim, Sekeun and Hu, Jiang and Liu, Zhengliang and Sun, Lichao and Li, Xiang and others},
  journal={Medical Image Analysis},
  volume={98},
  pages={103310},
  year={2024},
  publisher={Elsevier}
}

@article{wu2025medical,
  title={Medical sam adapter: Adapting segment anything model for medical image segmentation},
  author={Wu, Junde and Wang, Ziyue and Hong, Mingxuan and Ji, Wei and Fu, Huazhu and Xu, Yanwu and Xu, Min and Jin, Yueming},
  journal={Medical image analysis},
  volume={102},
  pages={103547},
  year={2025},
  publisher={Elsevier}
}

@inproceedings{zhuang2025bio2vol,
  title={Bio2vol: Adapting 2d biomedical foundation models for volumetric medical image segmentation},
  author={Zhuang, Jiaxin and Wu, Linshan and Ni, Xuefeng and Wang, Xi and Wang, Liansheng and Chen, Hao},
  booktitle={International Conference on Medical Image Computing and Computer-Assisted Intervention},
  pages={24--34},
  year={2025},
  organization={Springer}
}

@article{yang2025sam2,
  title={SAM2-3dMed: Empowering SAM2 for 3D Medical Image Segmentation},
  author={Yang, Yeqing and Xu, Le and Tian, Lixia},
  journal={arXiv preprint arXiv:2510.08967},
  year={2025}
}

@article{zhu2024medical,
  title={Medical sam 2: Segment medical images as video via segment anything model 2},
  author={Zhu, Jiayuan and Hamdi, Abdullah and Qi, Yunli and Jin, Yueming and Wu, Junde},
  journal={arXiv preprint arXiv:2408.00874},
  year={2024}
}

@article{gong20233dsam,
  title={3dsam-adapter: Holistic adaptation of sam from 2d to 3d for promptable tumor segmentation},
  author={Gong, Shizhan and Zhong, Yuan and Ma, Wenao and Li, Jinpeng and Wang, Zhao and Zhang, Jingyang and Heng, Pheng-Ann and Dou, Qi},
  journal={Medical Image Analysis},
  volume={98},
  pages={103324},
  year={2024},
  publisher={Elsevier}
}

@article{heller2021state,
  title={The state of the art in kidney and kidney tumor segmentation in contrast-enhanced CT imaging: Results of the KiTS19 challenge},
  author={Heller, Nicholas and Isensee, Fabian and Maier-Hein, Klaus H and Hou, Xiaoshuai and Xie, Chunmei and Li, Fengyi and Nan, Yang and Mu, Guangrui and Lin, Zhiyong and Han, Miofei and others},
  journal={Medical image analysis},
  volume={67},
  pages={101821},
  year={2021},
  publisher={Elsevier}
}

@article{ulman2017objective,
  title={An objective comparison of cell-tracking algorithms},
  author={Ulman, Vladim{\'\i}r and Ma{\v{s}}ka, Martin and Magnusson, Klas EG and Ronneberger, Olaf and Haubold, Carsten and Harder, Nathalie and Matula, Pavel and Matula, Petr and Svoboda, David and Radojevic, Miroslav and others},
  journal={Nature methods},
  volume={14},
  number={12},
  pages={1141--1152},
  year={2017},
  publisher={Nature Publishing Group UK London}
}

@article{hu2022lora,
  title={Lora: Low-rank adaptation of large language models.},
  author={Hu, Edward J and Shen, Yelong and Wallis, Phillip and Allen-Zhu, Zeyuan and Li, Yuanzhi and Wang, Shean and Wang, Lu and Chen, Weizhu and others},
  journal={ICLR},
  volume={1},
  number={2},
  pages={3},
  year={2022}
}

@inproceedings{lester2021prompt,
  title={The power of scale for parameter-efficient prompt tuning},
  author={Lester, Brian and Al-Rfou, Rami and Constant, Noah},
  booktitle={Proceedings of the 2021 conference on empirical methods in natural language processing},
  pages={3045--3059},
  year={2021}
}

@inproceedings{jia2022vpt,
  title={Visual prompt tuning},
  author={Jia, Menglin and Tang, Luming and Chen, Bor-Chun and Cardie, Claire and Belongie, Serge and Hariharan, Bharath and Lim, Ser-Nam},
  booktitle={European conference on computer vision},
  pages={709--727},
  year={2022},
  organization={Springer}
}

@article{dettmers2023qlora,
  title={Qlora: Efficient finetuning of quantized llms},
  author={Dettmers, Tim and Pagnoni, Artidoro and Holtzman, Ari and Zettlemoyer, Luke},
  journal={Advances in neural information processing systems},
  volume={36},
  pages={10088--10115},
  year={2023}
}

@inproceedings{liu2024dora,
  title={Dora: Weight-decomposed low-rank adaptation},
  author={Liu, Shih-Yang and Wang, Chien-Yi and Yin, Hongxu and Molchanov, Pavlo and Wang, Yu-Chiang Frank and Cheng, Kwang-Ting and Chen, Min-Hung},
  booktitle={Forty-first International Conference on Machine Learning},
  year={2024}
}

@article{zhang2023adalora,
  title={Adalora: Adaptive budget allocation for parameter-efficient fine-tuning},
  author={Zhang, Qingru and Chen, Minshuo and Bukharin, Alexander and Karampatziakis, Nikos and He, Pengcheng and Cheng, Yu and Chen, Weizhu and Zhao, Tuo},
  journal={arXiv preprint arXiv:2303.10512},
  year={2023}
}

@inproceedings{wang2021patch,
  title={Patch-free 3D medical image segmentation driven by super-resolution technique and self-supervised guidance},
  author={Wang, Hongyi and Lin, Lanfen and Hu, Hongjie and Chen, Qingqing and Li, Yinhao and Iwamoto, Yutaro and Han, Xian-Hua and Chen, Yen-Wei and Tong, Ruofeng},
  booktitle={International conference on medical image computing and computer-assisted intervention},
  pages={131--141},
  year={2021},
  organization={Springer}
}

@inproceedings{jeon2025no,
  title={No more sliding window: efficient 3D medical image segmentation with differentiable top-k patch sampling},
  author={Jeon, Young Seok and Yang, Hongfei and Fu, Huazhu and Kway, Yeshe and Feng, Mengling},
  booktitle={International Conference on Medical Image Computing and Computer-Assisted Intervention},
  pages={376--386},
  year={2025},
  organization={Springer}
}

@article{zadouri2023pushing,
  title={Pushing mixture of experts to the limit: Extremely parameter efficient moe for instruction tuning},
  author={Zadouri, Ted and {\"U}st{\"u}n, Ahmet and Ahmadian, Arash and Ermi{\c{s}}, Beyza and Locatelli, Acyr and Hooker, Sara},
  journal={arXiv preprint arXiv:2309.05444},
  year={2023}
}

@inproceedings{lin2019tsm,
  title={Tsm: Temporal shift module for efficient video understanding},
  author={Lin, Ji and Gan, Chuang and Han, Song},
  booktitle={IEEE Conf. Comput. Vis. Pattern Recog.},
  pages={7083--7093},
  year={2019}
}

@inproceedings{ryali2023hiera,
  title={Hiera: A hierarchical vision transformer without the bells-and-whistles},
  author={Ryali, Chaitanya and Hu, Yuan-Ting and Bolya, Daniel and Wei, Chen and Fan, Haoqi and Huang, Po-Yao and Aggarwal, Vaibhav and Chowdhury, Arkabandhu and Poursaeed, Omid and Hoffman, Judy and others},
  booktitle={International conference on machine learning},
  pages={29441--29454},
  year={2023},
  organization={PMLR}
}

@article{carion2025sam,
  title={Sam 3: Segment anything with concepts},
  author={Carion, Nicolas and Gustafson, Laura and Hu, Yuan-Ting and Debnath, Shoubhik and Hu, Ronghang and Suris, Didac and Ryali, Chaitanya and Alwala, Kalyan Vasudev and Khedr, Haitham and Huang, Andrew and others},
  journal={arXiv preprint arXiv:2511.16719},
  year={2025}
}

@article{dosovitskiy2020image,
  title={An image is worth 16x16 words: Transformers for image recognition at scale},
  author={Dosovitskiy, Alexey and Beyer, Lucas and Kolesnikov, Alexander and Weissenborn, Dirk and Zhai, Xiaohua and Unterthiner, Thomas and Dehghani, Mostafa and Minderer, Matthias and Heigold, Georg and Gelly, Sylvain and others},
  journal={arXiv preprint arXiv:2010.11929},
  year={2020}
}

@article{zhang2023customized,
  title={Customized segment anything model for medical image segmentation},
  author={Zhang, Kaidong and Liu, Dong},
  journal={arXiv preprint arXiv:2304.13785},
  year={2023}
}

@inproceedings{hu2024lga,
  title={Lga: A language guide adapter for advancing the sam model’s capabilities in medical image segmentation},
  author={Hu, Jihong and Li, Yinhao and Sun, Hao and Song, Yu and Zhang, Chujie and Lin, Lanfen and Chen, Yen-Wei},
  booktitle={International Conference on Medical Image Computing and Computer-Assisted Intervention},
  pages={610--620},
  year={2024},
  organization={Springer}
}

@article{wu2024mixture,
  title={Mixture of lora experts},
  author={Wu, Xun and Huang, Shaohan and Wei, Furu},
  journal={arXiv preprint arXiv:2404.13628},
  year={2024}
}

@article{li2024mixlora,
  title={Mixlora: Enhancing large language models fine-tuning with lora-based mixture of experts},
  author={Li, Dengchun and Ma, Yingzi and Wang, Naizheng and Ye, Zhengmao and Cheng, Zhiyuan and Tang, Yinghao and Zhang, Yan and Duan, Lei and Zuo, Jie and Yang, Cal and others},
  journal={arXiv preprint arXiv:2404.15159},
  year={2024}
}

@inproceedings{li2021prefix,
  title={Prefix-tuning: Optimizing continuous prompts for generation},
  author={Li, Xiang Lisa and Liang, Percy},
  booktitle={Proceedings of the 59th Annual Meeting of the Association for Computational Linguistics and the 11th International Joint Conference on Natural Language Processing (Volume 1: Long Papers)},
  pages={4582--4597},
  year={2021}
}

@inproceedings{houlsby2019parameter,
  title={Parameter-efficient transfer learning for NLP},
  author={Houlsby, Neil and Giurgiu, Andrei and Jastrzebski, Stanislaw and Morrone, Bruna and De Laroussilhe, Quentin and Gesmundo, Andrea and Attariyan, Mona and Gelly, Sylvain},
  booktitle={International conference on machine learning},
  pages={2790--2799},
  year={2019},
  organization={PMLR}
}

@article{chen2022adaptformer,
  title={Adaptformer: Adapting vision transformers for scalable visual recognition},
  author={Chen, Shoufa and Ge, Chongjian and Tong, Zhan and Wang, Jiangliu and Song, Yibing and Wang, Jue and Luo, Ping},
  journal={Advances in Neural Information Processing Systems},
  volume={35},
  pages={16664--16678},
  year={2022}
}

@article{antonelli2022medical,
  title={The medical segmentation decathlon},
  author={Antonelli, Michela and Reinke, Annika and Bakas, Spyridon and Farahani, Keyvan and Kopp-Schneider, Annette and Landman, Bennett A and Litjens, Geert and Menze, Bjoern and Ronneberger, Olaf and Summers, Ronald M and others},
  journal={Nature communications},
  volume={13},
  number={1},
  pages={4128},
  year={2022},
  publisher={Nature Publishing Group UK London}
}

@article{bilic2023liver,
  title={The liver tumor segmentation benchmark (lits)},
  author={Bilic, Patrick and Christ, Patrick and Li, Hongwei Bran and Vorontsov, Eugene and Ben-Cohen, Avi and Kaissis, Georgios and Szeskin, Adi and Jacobs, Colin and Mamani, Gabriel Efrain Humpire and Chartrand, Gabriel and others},
  journal={Medical image analysis},
  volume={84},
  pages={102680},
  year={2023},
  publisher={Elsevier}
}

@article{svoboda2016mitogen,
  title={MitoGen: a framework for generating 3D synthetic time-lapse sequences of cell populations in fluorescence microscopy},
  author={Svoboda, David and Ulman, Vladimir},
  journal={IEEE transactions on medical imaging},
  volume={36},
  number={1},
  pages={310--321},
  year={2016},
  publisher={IEEE}
}

@inproceedings{wang2021transbts,
  title={Transbts: Multimodal brain tumor segmentation using transformer},
  author={Wang, Wenxuan and Chen, Chen and Ding, Meng and Yu, Hong and Zha, Sen and Li, Jiangyun},
  booktitle={International conference on medical image computing and computer-assisted intervention},
  pages={109--119},
  year={2021},
  organization={Springer}
}

@article{zhou2021nnformer,
  title={nnformer: Interleaved transformer for volumetric segmentation},
  author={Zhou, Hong-Yu and Guo, Jiansen and Zhang, Yinghao and Yu, Lequan and Wang, Liansheng and Yu, Yizhou},
  journal={arXiv preprint arXiv:2109.03201},
  year={2021}
}

@inproceedings{tang2022self,
  title={Self-supervised pre-training of swin transformers for 3d medical image analysis},
  author={Tang, Yucheng and Yang, Dong and Li, Wenqi and Roth, Holger R and Landman, Bennett and Xu, Daguang and Nath, Vishwesh and Hatamizadeh, Ali},
  booktitle={Proceedings of the IEEE/CVF conference on computer vision and pattern recognition},
  pages={20730--20740},
  year={2022}
}

@article{shaker2024unetrpp,
  title={UNETR++: delving into efficient and accurate 3D medical image segmentation},
  author={Shaker, Abdelrahman and Maaz, Muhammad and Rasheed, Hanoona and Khan, Salman and Yang, Ming-Hsuan and Khan, Fahad Shahbaz},
  journal={IEEE Transactions on Medical Imaging},
  volume={43},
  number={9},
  pages={3377--3390},
  year={2024},
  publisher={IEEE}
}

@article{lee20223d,
  title={3d ux-net: A large kernel volumetric convnet modernizing hierarchical transformer for medical image segmentation},
  author={Lee, Ho Hin and Bao, Shunxing and Huo, Yuankai and Landman, Bennett A},
  journal={arXiv preprint arXiv:2209.15076},
  year={2022}
}

@inproceedings{xu2016deep,
  title={Deep interactive object selection},
  author={Xu, Ning and Price, Brian and Cohen, Scott and Yang, Jimei and Huang, Thomas S},
  booktitle={Proceedings of the IEEE conference on computer vision and pattern recognition},
  pages={373--381},
  year={2016}
}

@article{luo2021mideepseg,
  title={MIDeepSeg: Minimally interactive segmentation of unseen objects from medical images using deep learning},
  author={Luo, Xiangde and Wang, Guotai and Song, Tao and Zhang, Jingyang and Aertsen, Michael and Deprest, Jan and Ourselin, Sebastien and Vercauteren, Tom and Zhang, Shaoting},
  journal={Medical image analysis},
  volume={72},
  pages={102102},
  year={2021},
  publisher={Elsevier}
}

@inproceedings{zhou2023interactive,
  title={Interactive segmentation as gaussion process classification},
  author={Zhou, Minghao and Wang, Hong and Zhao, Qian and Li, Yuexiang and Huang, Yawen and Meng, Deyu and Zheng, Yefeng},
  booktitle={Proceedings of the IEEE/CVF conference on computer vision and pattern recognition},
  pages={19488--19497},
  year={2023}
}

@article{zou2023segment,
  title={Segment everything everywhere all at once},
  author={Zou, Xueyan and Yang, Jianwei and Zhang, Hao and Li, Feng and Li, Linjie and Wang, Jianfeng and Wang, Lijuan and Gao, Jianfeng and Lee, Yong Jae},
  journal={Advances in neural information processing systems},
  volume={36},
  pages={19769--19782},
  year={2023}
}

@inproceedings{ben2022graph,
  title={Graph neural network for cell tracking in microscopy videos},
  author={Ben-Haim, Tal and Raviv, Tammy Riklin},
  booktitle={European Conference on Computer Vision},
  pages={610--626},
  year={2022},
  organization={Springer}
}

@inproceedings{bragantini2024ucmtracking,
  title={Large-scale multi-hypotheses cell tracking using ultrametric contours maps},
  author={Bragantini, Jord{\~a}o and Lange, Merlin and Royer, Lo{\"\i}c},
  booktitle={European Conference on Computer Vision},
  pages={36--54},
  year={2024},
  organization={Springer}
}

@article{bragantini2024ultrack,
  title={Ultrack: pushing the limits of cell tracking across biological scales},
  author={Bragantini, Jord{~a}o and Theodoro, Ilan and Zhao, Xiang and Huijben, Teun APM and Hirata-Miyasaki, Eduardo and VijayKumar, Shruthi and Balasubramanian, Akilandeswari and Lao, Tiger and Agrawal, Richa and Xiao, Sheng and others},
  journal={bioRxiv},
  pages={2024--09},
  year={2024},
  publisher={Cold Spring Harbor Laboratory}
}

@inproceedings{gallusser2024trackastra,
  title={Trackastra: Transformer-based cell tracking for live-cell microscopy},
  author={Gallusser, Benjamin and Weigert, Martin},
  booktitle={European conference on computer vision},
  pages={467--484},
  year={2024},
  organization={Springer}
}

@inproceedings{chen2025segment,
  title={Segment Anything for Cell Tracking},
  author={Chen, Zhu and Edg{\"u}, Mert and Jin, Er and Stegmaier, Johannes},
  booktitle={International Workshop on Foundation Models for General Medical AI},
  pages={12--22},
  year={2025},
  organization={Springer}
}

@article{stringer2021cellpose,
  title={Cellpose: a generalist algorithm for cellular segmentation},
  author={Stringer, Carsen and Wang, Tim and Michaelos, Michalis and Pachitariu, Marius},
  journal={Nature methods},
  volume={18},
  number={1},
  pages={100--106},
  year={2021},
  publisher={Nature Publishing Group US New York}
}

\clearpage

\appendix
\renewcommand{\thesection}{\Alph{section}}

\section{Decoder Architectures}
\label{sup:decoders}

\begin{figure}[t]
  \centering
  \includegraphics[width=\linewidth]{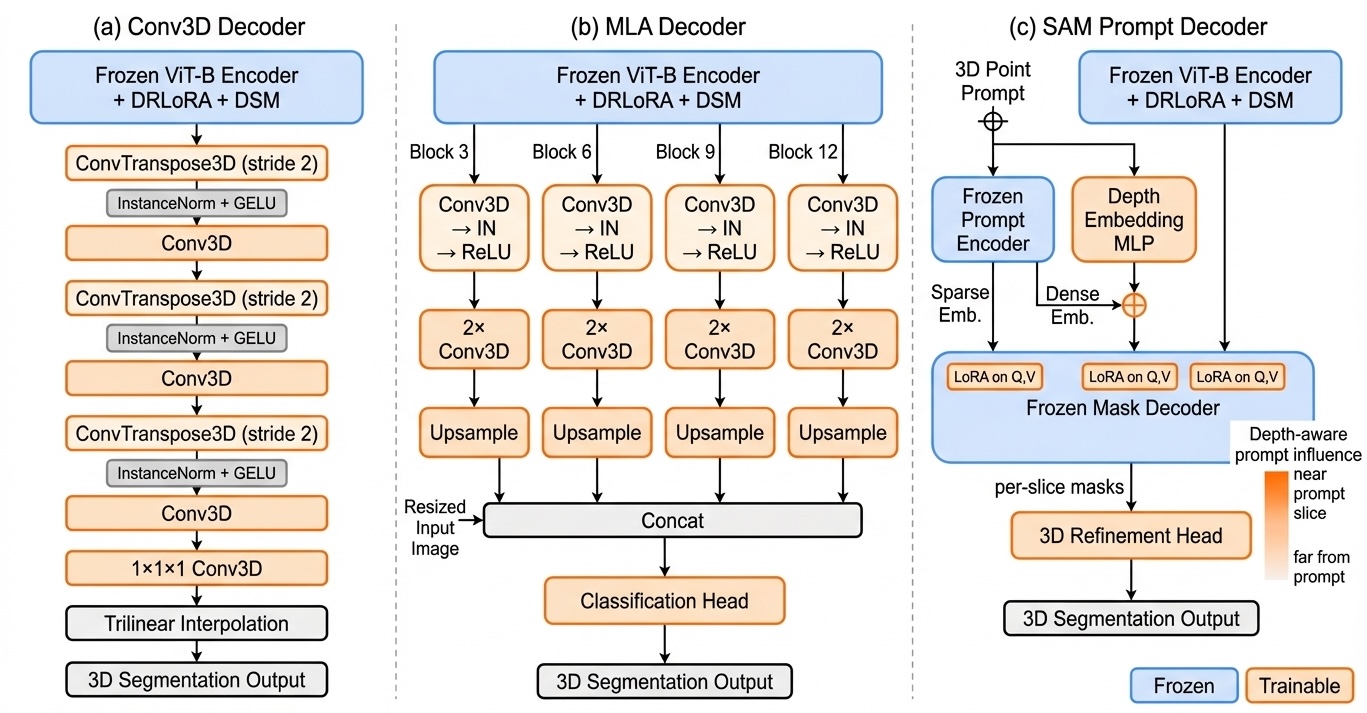}
  \caption{%
  \textbf{Three decoder architectures for SAM+D.}
  (a)~Conv3D decoder: directly upsamples the 3D feature volume
  through transposed convolutions.
  (b)~MLA decoder: aggregates intermediate features from encoder
  blocks $\{3, 6, 9, 12\}$ through parallel pathways.
  (c)~SAM prompt decoder (default): reuses SAM's frozen prompt encoder and mask decoder with LoRA injection and a learnable depth embedding.
  Orange: trainable; blue: frozen; gray: zero-parameter 
}
\label{fig:decoders}
\end{figure}

To evaluate the effectiveness of our DRLoRA
and DSM modules independently of decoder choice, we design three decoder
architectures with varying complexity and inductive biases, as illustrated in
Fig.~\ref{fig:decoders}. All
three achieve competitive performance (Table~1 in the main paper),
confirming that the gains stem primarily from the encoder adaptation
rather than a specific decoder design.

All three decoders share the same frozen SAM ViT-B encoder augmented
with DRLoRA and DSM. The encoder processes
them through 12 transformer blocks. Features pass through
SAM's frozen neck (two $256$-channel convolutions with LayerNorm),
yielding a feature volume of shape $(B, 256, 32, 32, 32)$. The three
decoders differ only in how this volume is decoded into the final
segmentation.

Note that all three decoders require a single point prompt, but its
role differs across designs. For the Conv3D and MLA decoders, the
point prompt serves solely as a spatial prior for cropping a
region of interest (ROI) from the full volume; it does not
participate in the decoding process itself. For the SAM prompt
decoder, the point additionally enters SAM's prompt encoder to
guide mask generation. In all cases, the cropped volume is processed
in a single forward pass without sliding window inference---our
method remains fully patch-free regardless of decoder choice.

\subsection{Decoder 1: Conv3D.}
This decoder directly upsamples the 3D feature volume through two
transposed convolution stages. Each stage consists of a
ConvTranspose3d (kernel $2$, stride $2$) followed by InstanceNorm3d,
GELU, a Conv3d ($3\times3\times3$, padding $1$), InstanceNorm3d, and
GELU. The first stage maps $256\!\to\!128$ channels at
$64^3$ resolution; the second maps $128\!\to\!64$ channels at
$128^3$ resolution. A final Conv3d ($1\times1\times1$) projects to
$2$ output classes, followed by trilinear interpolation to the
target volume size. This decoder adds 881,154 trainable parameters,
bringing the total to 3,286,626 (3.7\% of 89.77\,M).

\subsection{Decoder 2: Multi-Layer Aggregation (MLA).}
This decoder captures intermediate features from encoder blocks
$\{3, 6, 9, 12\}$ (0-indexed: 2, 5, 8, 11). Each feature map
$(B, 768, 32, 32, 32)$ is first projected to 256 channels via a
Conv3d ($1\times1\times1$) with InstanceNorm3d and ReLU
(786,432 params total for four projections). Each projected feature
then passes through a two-layer pathway: Conv3d
($256\!\to\!128$, $3\times3\times3$) $\to$ InstanceNorm3d $\to$ ReLU
$\to$ Conv3d ($128\!\to\!128$, $3\times3\times3$) $\to$
InstanceNorm3d $\to$ ReLU $\to$ trilinear interpolation to the
target size (5,308,416 params for four pathways). The four
128-channel outputs are concatenated into a 512-channel volume, which
is concatenated with the trilinearly resized input image (1 channel)
to form a 513-channel tensor. A classification head consisting of
Conv3d ($513\!\to\!128$, $3\times3\times3$) $\to$ InstanceNorm3d
$\to$ ReLU $\to$ Conv3d ($128\!\to\!2$, $1\times1\times1$) produces
the final output. The decoder adds 7,868,034 trainable parameters,
totalling 10,273,506 (10.6\% of 96.76\,M).

\subsection{Decoder 3: SAM Prompt Decoder (default).}
This decoder reuses SAM's original prompt encoder and mask decoder
with minimal modifications. The frozen prompt encoder produces sparse
embeddings from point prompts and dense positional embeddings at
$32\times32$ resolution. A trainable depth embedding MLP
(Linear($1\!\to\!256$) $\to$ GELU $\to$ Linear($256\!\to\!256$);
66,304 params) encodes the relative $z$-distance between each slice
and the prompt point, and adds it to the dense prompt embeddings.

SAM's mask decoder consists of a two-layer two way Transformer
followed by an upscaling pathway. We inject LoRA (rank $r\!=\!16$)
into the Q and V projections of all attention sub-layers: self
attention, cross attention (token-to-image), and cross attention
(image-to-token) in each of the two transformer blocks, plus the
final token-to-image attention layer, totalling 14 Q/V LoRA pairs
(94,208 params). All LayerNorms in the decoder are unfrozen
(4,096 params). The remaining components---output upscaling
convolutions, hypernetwork MLPs, and IoU prediction head---stay
frozen. Per-slice mask logits are reshaped to
$(B, 1, 32, 128, 128)$ and passed through a lightweight 3D
refinement head: Conv3d ($1\!\to\!32$, $3\times3\times3$) $\to$
InstanceNorm3d $\to$ ReLU $\to$ Conv3d ($32\!\to\!2$,
$1\times1\times1$), adding 962 params for cross-slice smoothing.
The output is trilinearly interpolated to the target volume size.

Table~\ref{tab:sup_decoder_params} summarises the trainable parameter
breakdown.

\begin{table}[h]
\centering
\caption{Trainable parameter breakdown for the SAM prompt decoder
(default). }
\label{tab:sup_decoder_params}
\begin{tabular}{lr}
\toprule
\textbf{Component} & \textbf{\#Params} \\
\midrule
Encoder DRLoRA (12 blocks $\times$ 2 Q/V) & 2,368,608 \\
Encoder LayerNorms (12 blocks) & 36,864 \\
\midrule
Decoder LoRA (14 Q/V pairs, $r\!=\!16$) & 94,208 \\
Depth embedding MLP & 66,304 \\
Decoder LayerNorms & 4,096 \\
3D refinement head & 962 \\
\midrule
\textbf{Total trainable} & \textbf{2,571,042} \\
\textbf{Trainable ratio} & 2.8\% \\
\bottomrule
\end{tabular}
\end{table}

\section{4D Inference Pipeline}
\label{sup:4d_inference}

We describe the two evaluation modes for 4D spatiotemporal
segmentation and tracking on Fluo-N3DH-SIM+.

\subsection{Mode 1: GT Segmentation + SAM2 Linking}

In this mode the ground-truth masks are known at every frame, so segmentation is perfect and only the cross-frame linking is evaluated. We track one cell at a time: the cell is initialised at the first frame of each clip with a point prompt at its GT-mask centroid, and SAM2+D propagates it forward through the clip via DSM and
LoRA-adapted memory attention, which yields ranked candidate matches in the
following frame(s). The clip length $T$ is flexible; the minimal case $T\!=\!2$ corresponds to a single frame-to-frame link. Because all masks are known, we use them to validate each link, rejecting matches to non-existent cells and falling back to the next-highest-probability candidate. The model outputs binary masks for the whole clip, and this mode therefore isolates linking quality from segmentation.

\subsection{Mode 2: Cellpose Detector + SAM2 Propagation}

This mode represents a practical pipeline where no ground truth is
available. It consists of three stages.

\subsubsection{Stage 1: Detection and association.}
Cellpose produces a 3D instance segmentation for every frame. We first
filter these detections by volume, discarding instances outside a
plausible cell-size range (min/max). We then associate detections into
tracks: consecutive frames are linked by the Hungarian algorithm on centroid distances, giving a one-to-one assignment between frame $t$ and $t{+}1$. Finally, we handle track lifecycle: a track left unmatched for more than three consecutive frames is terminated, and a division is declared when an unmatched detection appears adjacent to an active track whose volume has dropped to $\leq$70\% of its previous value (a mother cell splitting into daughters).

\subsubsection{Stage 2: SAM2+D propagation.}
Each tracked cell is then re-segmented by propagation rather than trusting Cellpose alone. SAM2+D is prompted with the cell's centroid at its first detection and propagates the mask forward over a 16-frame temporal window using its memory bank.
SAM2+D is the primary segmentation authority for established tracks, while Cellpose is still consulted at every frame to discover new cells and divisions. To decide between the two sources per frame, we use SAM2's native predicted-IoU score as a confidence estimate: high-confidence SAM2+D masks are kept, while low-confidence frames fall back to the Cellpose segmentation. Two consistency checks guard against drift---IoU with the previous mask $>0.2$ and a volume ratio $>0.3$---ensuring spatial coherence across frames.

\subsubsection{Stage 3: Re-association.}
Finally, because Stage 2 has changed the masks, we recompute the
track-to-detection correspondences on the refined masks and emit the result in CTC format. Table~\ref{tab:sup_4d_infer} lists the inference hyperparameters.

\begin{table}[htbp]
\centering
\caption{4D inference hyperparameters.}
\label{tab:sup_4d_infer}
\begin{tabular}{ll}
\toprule
\textbf{Hyperparameter} & \textbf{Value} \\
\midrule
\multicolumn{2}{l}{\textit{SAM2+D inference}} \\
Temporal window & $T\!=\!16$ frames \\
Prompt type & Point (centroid of last known mask) \\
ROI padding factor & $2.0\times$ \\
ROI min size & $(16, 64, 64)$ \\
Spatial resolution & $(32, 256, 256)$ \\
Precision & BF16 \\
\midrule
\multicolumn{2}{l}{\textit{Association}} \\
Centroid match distance & 50 voxels \\
Min mask volume & 100 voxels \\
Max dormant frames & 3 \\
Dormant match distance & 80 voxels \\
\midrule
\multicolumn{2}{l}{\textit{Division detection}} \\
Parent--daughter distance & 50 voxels \\
Volume shrink threshold & $\leq$70\% \\
\midrule
\multicolumn{2}{l}{\textit{SAM2+D consistency checks}} \\
IoU with previous mask & $>0.2$ \\
Volume ratio & $>0.3$ \\
\bottomrule
\end{tabular}
\end{table}

\section{Training Hyperparameter Settings}
\label{sup:hyperparams}

\subsection{3D Setting}
\label{sup:hyper_3d}

All 3D experiments use SAM ViT-B as the frozen backbone with DRLoRA
and DSM inserted into each of the 12 encoder blocks.
Table~\ref{tab:sup_3d_hyper} provides the complete hyperparameter
configuration.

\begin{table}[htbp]
\centering
\caption{Hyperparameters for the 3D setting.}
\label{tab:sup_3d_hyper}
\begin{tabular}{ll}
\toprule
\textbf{Hyperparameter} & \textbf{Value} \\
\midrule
\multicolumn{2}{l}{\textit{Model}} \\
Backbone & SAM ViT-B (frozen) \\
DRLoRA rank / experts & $r\!=\!16$, $E\!=\!4$ (Q and V, 12 blocks) \\
DSM shift ratio ($\alpha$) & 0.25 (depth axis) \\
Decoder LoRA (SAM dec.) & rank 16, 14 Q/V pairs \\
Depth embedding & $1\!\to\!256\!\to\!256$ (MLP) \\
Output classes & 2 \\
\midrule
\multicolumn{2}{l}{\textit{Training}} \\
Optimizer & AdamW (weight decay $10^{-4}$) \\
Learning rate & $4\times10^{-4}$  \\
Epochs & 500 \\
Batch size & 3 per GPU \\
Loss & DiceCE \\
Precision & AMP (float16) \\
\midrule
\multicolumn{2}{l}{\textit{Data}} \\
Crop size & $128\times128\times128$ \\
Depth slices ($D_s$) / spatial res. & 32 / $512\times512$ \\
Target spacing & $1\times1\times1$\,mm (isotropic) \\
Point prompts & 1 per volume \\
\midrule
\multicolumn{2}{l}{\textit{Per-dataset intensity normalisation}} \\
KiTS & clip $[-54, 247]$, $\mu\!=\!59.54$, $\sigma\!=\!55.46$ \\
LiTS & clip $[-48, 163]$, $\mu\!=\!60.06$, $\sigma\!=\!40.20$ \\
Pancreas & clip $[-39, 204]$, $\mu\!=\!68.45$, $\sigma\!=\!63.42$ \\
Colon & clip $[-57, 175]$, $\mu\!=\!65.18$, $\sigma\!=\!32.65$ \\
\midrule
\multicolumn{2}{l}{\textit{Augmentation (training only)}} \\
RandZoom (KiTS, LiTS) & $p\!=\!0.8$, scale $[0.85, 1.25]$ \\
RandRotate (Pancreas, Colon) & $p\!=\!0.3$, range $30°$ \\
RandFlip & $p\!=\!0.5$, each axis \\
RandRotate90 & $p\!=\!0.5$, up to 3 rotations \\
\bottomrule
\end{tabular}
\end{table}

\subsection{4D Setting}
\label{sup:hyper_4d}

All 4D experiments use SAM2.1 Hiera Base+ as the frozen backbone
with DRLoRA, DSM, and memory attention LoRA.
Table~\ref{tab:sup_4d_hyper} provides the complete hyperparameter
configuration.

\begin{table}[h]
\centering
\caption{Hyperparameters for the 4D setting.}
\label{tab:sup_4d_hyper}
\begin{tabular}{ll}
\toprule
\textbf{Hyperparameter} & \textbf{Value} \\
\midrule
\multicolumn{2}{l}{\textit{Model}} \\
Backbone & SAM2.1 Hiera Base+ (frozen) \\
DRLoRA rank / experts & $r\!=\!16$, $E\!=\!4$ \\
DSM shift ratio ($\alpha$) & 0.25 (depth axis) \\
Memory attention LoRA & rank 16, all 4 layers, Q and V \\
Output classes & 2 \\
\midrule
\multicolumn{2}{l}{\textit{Training}} \\
Optimizer & AdamW (weight decay $0.01$) \\
Learning rate & $1\times10^{-4}$ \\
LR schedule & 5-epoch linear warmup $\to$ cosine
              ($\eta_{\min}\!=\!10^{-6}$) \\
Epochs & 500 \\
Batch size & 2 per GPU $\times$ 4 accum.\ steps (eff.\ 8) \\
Gradient clipping & Max norm 1.0 \\
Precision & BF16 \\
Loss & DiceCE ($\lambda_{\text{dice}}\!=\!\lambda_{\text{ce}}\!=\!1.0$) \\
Hardware & $3\times$ NVIDIA RTX 6000 Pro \\
\midrule
\multicolumn{2}{l}{\textit{Data}} \\
Dataset & Fluo-N3DH-SIM+ (train: Seq01, test: Seq02) \\
Clip length / stride & 16 frames / stride 2 \\
Spatial size & $32\times256\times256$ ($D\!\times\!H\!\times\!W$) \\
Augmentation & RandFlip $p\!=\!0.5$ per axis (W, H, D) \\
\bottomrule

\end{tabular}
\end{table}

\section{Additional Comparisons}

\subsection{Cross-Slice Operator Comparison}
Table~\ref{tab:crossslice} compares DSM against a parameter-matched cross-slice
attention and a Conv3D adapter on LiTS; DSM is most accurate while adding zero
per-block parameters.

\subsection{SAM-Based Methods}
Table~\ref{tab:public} reports the SAM-based methods omitted from the main
comparison (their pretraining corpora overlap our public test splits), together with a contamination-free private liver-cancer CT dataset
(train/val/test\,$=$\,86/13/24). SAM+D leads on the private data while training the fewest parameters.

The private dataset was provided by the Department of Radiology, Sir Run Run Shaw Hospital, using non-contrast (NC) phase CT; all scans were fully de-identified and used solely for research in accordance with the institution's ethical guidelines.

\subsection{Depth Routing vs.\ Content-Based MoE-LoRA}
Table~\ref{tab:moelora} compares DRLoRA against existing MoE-LoRA routers on LiTS under a matched parameter budget.

\begin{table}[t]
\centering
\scriptsize
\setlength{\tabcolsep}{3pt}
\renewcommand{\arraystretch}{1.05}
\caption{\textbf{Cross-slice operators on LiTS} at matched per-block budget. \#TP: full SAM+D trainable parameters.}
\label{tab:crossslice}
\begin{tabular}{l|ccc|c}
\toprule
 & Op & Per-block & \#TP & Dice/NSD \\
\midrule
\textbf{DSM, ours} & shift ($\alpha{=}0.25$) & 0 & 2.57M & \textbf{63.33/73.52} \\
Cross-slice attention        & MHA along $D$ ($h{=}4$)  & 115K & 3.96M & 55.01/66.20 \\
Conv3D adapter      & $k{=}3$ + GN+GELU       & 113K & 3.93M & 53.62/62.32 \\
\bottomrule
\end{tabular}
\end{table}

\begin{table}[htbp]
\centering
\scriptsize
\setlength{\tabcolsep}{1.5pt}
\renewcommand{\arraystretch}{1.05}
\caption{\textbf{Comparison with SAM-based SOTA methods.} Each cell reports Dice/NSD (\%). Top: four public CT benchmarks; \textcolor{contamred}{\textbf{Yes}}\textcolor{contamred}{: test set used in pretraining (not a fair comparison)}. Bottom: contamination-free private liver-cancer CT data (train/val/test = 86/13/24). $^{\dagger}$SegVol uses text (e.g., ``a computerized tomography of a kidney tumor'')$+$points$+$zoom-in. \#TP: trainable parameters.}
\begin{tabular}{l|>{\columncolor{gray!18}}c>{\columncolor{gray!18}}c>{\columncolor{gray!18}}c|ccc}
\toprule
& \cellcolor{white}\shortstack{SAM-Med3D\\1pt/vol} & \cellcolor{white}\shortstack{SegVol$^{\dagger}$\\7pt+1txt/vol} & \cellcolor{white}\shortstack{MedSAM2\\1bbx/vol} & \shortstack{Med-SA\\1pt/slice} & \shortstack{MA-SAM\\no prompt} & \shortstack{SAM+D\\1pt/vol} \\
\midrule
\textbf{\shortstack{Test set used\\pretraining?}} &
\textcolor{contamred}{\textbf{Yes}} & \textcolor{contamred}{\textbf{Yes}} & \textcolor{contamred}{\textbf{Yes}} & No & No & No \\
\midrule
KiTS              & 81.6/92.6 & 52.0/66.3 & 80.2/87.2 & 70.7/82.4 & 69.3/65.8 & \textbf{84.7/92.1} \\
LiTS              & 46.9/61.8 & 67.6/80.4 & 37.4/52.9 & 51.6/72.0 & 56.3/62.9 & \textbf{63.3/73.5} \\
Pancreas          & 64.7/90.6 & 69.3/91.2 & 40.5/52.1 & 43.6/73.4 & 34.7/47.7 & \textbf{59.7/79.1} \\
Colon             & 67.9/93.3 & 73.7/91.5 & 61.9/75.9 & 53.6/74.9 & 50.3/56.4 & \textbf{63.7/77.7} \\
\midrule
\midrule
\textbf{\shortstack{Test set used\\pretraining?}} &
\cellcolor{White}No & \cellcolor{White}No & \multicolumn{1}{c}{No} & \cellcolor{White}No & \cellcolor{White}No & \cellcolor{White}No \\
\midrule
Private liver & \cellcolor{White}59.1/77.4 & \cellcolor{White}55.9/63.7 & \multicolumn{1}{c}{73.2/81.4} & \cellcolor{White}54.7/63.4 & \cellcolor{White}52.5/56.4 & \textbf{75.3/82.5} \\
\midrule
\#TP & \cellcolor{white}100.5M & \cellcolor{white}146.1M & \multicolumn{1}{c}{39.0M} & 13.0M & 63.0M & \textbf{2.6M} \\
\bottomrule
\end{tabular}
\label{tab:public}
\end{table}

\begin{table}[htbp]
\centering
\scriptsize
\setlength{\tabcolsep}{2.5pt}
\renewcommand{\arraystretch}{1.05}
\caption{\textbf{DRLoRA vs.\ existing MoE-LoRA methods.} Last row: Dice/NSD (\%) on LiTS.}
\begin{tabular}{l|ccc|c}
\toprule
 & MoLoRA~[49] & MoLE~[46] & MixLoRA~[27] & \textbf{DRLoRA} \\
\midrule
Routing signal     & content & content & content & \textbf{depth $z$} \\
Aux.\ balance loss & ---     & importance & load-balance & \textbf{none} \\
Router collapse risk & yes   & mitigated  & mitigated    & \textbf{none} \\
Router params      & $\sim$3K & $\sim$6K & $\sim$3K & \textbf{$\sim$0.14K} \\
Target domain      & NLP & instruct.\ tuning & LLM FT & \textbf{3D imaging} \\
\midrule
Dice/NSD (LiTS)    & 55.47/62.69 & 56.10/64.76 & 53.78/62.47 & \textbf{63.33/73.52} \\
\bottomrule
\end{tabular}
\label{tab:moelora}
\end{table}

\end{document}